\documentclass[final]{cvpr}

\usepackage{times}
\usepackage{epsfig}
\usepackage{graphicx}
\usepackage{amsmath}
\usepackage{amssymb}
\usepackage{subfigure}

\usepackage{multirow}
\usepackage{graphicx}

\usepackage[pagebackref=true,breaklinks=true,colorlinks,bookmarks=false]{hyperref}

\def\cvprPaperID{****} 

\begin{document}


\title{The Selectivity and Competition of the Mind's Eye in Visual Perception}

\author{Edward Kim$^{1}$, Maryam Daniali$^{1}$, Jocelyn Rego$^{1}$, Garrett T. Kenyon$^2$\\
$^1$Department of Computer Science, Drexel University, PA\\
$^2$Los Alamos National Laboratory, Los Alamos, NM\\
{\tt\small ek826@drexel.edu,md3464@drexel.edu,jr3548@drexel.edu,gkeynon@lanl.gov}
}

\maketitle
\thispagestyle{empty}

\begin{abstract}
Research has shown that neurons within the brain are selective to certain stimuli.  For example, the fusiform face area (FFA) region is known by neuroscientists to selectively activate when people see faces over non-face objects.  However, the mechanisms by which the primary visual system directs information to the correct higher levels of the brain are currently unknown.   In our work, we mimic several high-level neural mechanisms of perception by creating a novel computational model that incorporates lateral and top down feedback in the form of hierarchical competition.  Not only do we show that these elements can help explain the information flow and selectivity of high level areas within the brain, we also demonstrate that these neural mechanisms provide the foundation of a novel classification framework that rivals traditional supervised learning in computer vision.   Additionally, we present both quantitative and qualitative results that demonstrate that our generative framework is consistent with neurological themes and enables simple, yet robust category level classification. 
\vspace{-0.5cm}
\end{abstract}

\section{Introduction}
\label{sec:intro}
\begin{figure}
    \centering
    \includegraphics[width=7.5cm]{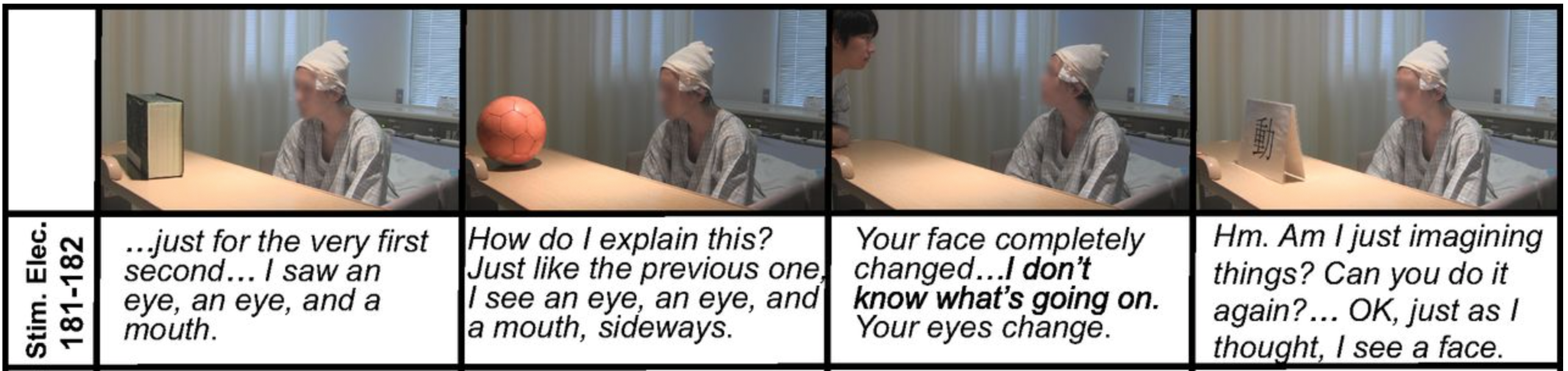}
    \caption{Experiment on seeing Facephenes from \cite{schalk2017facephenes}.  In our research, we build a multipath competitive framework that can help us understand this patient's experience, explain how information flows in visual perception, and inform a new kind of classification framework based on category competition.}
    \label{fig:facephenes}
    \vspace{-0.5cm}
\end{figure}

In 2017, a 26-year old patient at Asahikawa Medical University was being treated for intractable epilepsy.  This patient had subdural electrodes placed on a specific part of the brain that corresponds to what is known as the fusiform face area (FFA) \cite{schalk2017facephenes}.   When researchers artificially stimulated neurons in the FFA, the patient hallucinated faces or face parts in non-face, everyday objects.  As seen in Figure \ref{fig:facephenes}, the patient reports, ``I saw an eye, an eye, and a mouth.  I started thinking ‘what is this?’, but the next thing I noticed, I was just looking at this box.''

The FFA region is known by neuroscientists to selectively activate when people see faces, specifically upright faces, compared to the activations elicited by non-face objects \cite{kanwisher2010functional}.  Additional evidence exists revealing that the visual processing of faces and objects occur in different pathways \cite{griffin2019face}. Exploring deeper into the cerebral cortex, we can see that the FFA is one of many specialized, high level areas within the brain.  The FFA exists in the Inferior Temporal (IT) Cortex, the part of the brain critical for visual object memory and recognition, colloquially referred to as the \textit{Mind's Eye}.  Other specialized areas within the temporal cortex include selectivity for visual scenes or buildings (parahippocampal place area, PPA), for body parts (extrastriate body area, EBA), and for reading words (visual word form area, VWFA).  However, the mechanism by which the primary visual system directs information to the correct higher levels of the brain are currently unknown.  

\textit{How do the low-level, primary visual areas of the brain know where to send visual input information?}  This would imply that the low level areas have already done some sort of recognition of the input stimulus in order to route the information correctly to the higher levels in IT.   Some have hypothesized that there exists some low level gating mechanism that performs a rough detection and then forwards the information to specialized expert models, e.g. gating with a mixture of experts model \cite{masoudnia2014mixture, tsao2008mechanisms}.  Others have used a category template model that could roughly match an input stimulus.  A dot product between stimulus and template would measure similarity and then direct the stimulus down a certain pathway \cite{kay2017bottom}.  Others have built specialized hierarchical models specific for certain tasks that are pre-selected by a separate, high-level goal-driven model \cite{yamins2016using}.  

A shortcoming of all these previously described models is that they are implemented in a feed forward fashion.  We know that the brain is not strictly feed forward, but instead contains both lateral and top-down connections.  In fact, there are generally more feedback connections in the brain than there are feed forward \cite{cao2015look,kim2018deep}.  Purely feed forward models cannot capture the full account of the dynamics of neural systems \cite{yamins2016using}.  The gating or template model also fails to explain the reliable (yet diminished) object response and sensitivity to things like cars and planes within the FFA \cite{mcgugin2012high}.  Further, feed forward models cannot explain dynamic brain responses with identical stimuli since the responses are based solely as a function of the stimulus \cite{kay2017bottom}.

In our work, we develop a novel classification framework that rivals traditional supervised learning by mimicking neural mechanisms of perception. Specifically our contributions are,  1. We devise a novel algorithm and framework that incorporates lateral and top down feedback in the form of hierarchical competition that can explain the information flow and selectivity of high level areas within the brain.   2. We show that our model reflects general themes observed in the visual system and its responses are consistent with specific experiments in the neuroscience literature.   3. We show that our generative framework can perform category level image classification with superior performance to a comparable supervised convolutional neural network.  We demonstrate our results on the problem of face detection bias that has been uncovered in deep learning.  Given biased training data, our competitive multiscale model is able to overcome this bias given the nature of our holistic, coarse-to-fine processing approach.

\section{Background}
\subsection{Background in Visual Selectivity}
\textbf{Theme 1: Selectivity in the visual cortex can be achieved through competition.}
Research has shown that neurons within the brain are selective to certain stimuli.  Early work by Hubel and Wiesel demonstrated that cat V1 neurons were sensitive to the placement, orientation, and direction of movement of oriented edges \cite{hubel1962receptive}.  Neurons at this level compete to represent stimuli as evidenced by the mutual suppression of their visually evoked responses in their receptive field \cite{kastner2001neural}.   This surround suppression can be  explained by a local competitive mechanisms operating within a cortical column \cite{you2016surround}.  We see similar patterns of stimulus selectivity at higher levels of the brain i.e. in the inferior temporal gyrus or IT, where regions are selective to specific objects, faces, body parts, places, and words.  Literature shows competition between these high level areas, e.g. words and faces, for cortical territory \cite{kubota2019word}.  Additionally, we know that these high-level areas are actively competing at time of inference since object grouping based upon canonical configurations have shown to reduce competitive interactions as fewer disparate objects need to compete for a representation \cite{kaiser2014object}.  Also neurons selective for different objects mutually inhibit each other in the presence of their preferred stimulus \cite{norman2012category}, evidenced by a measured reduced blood oxygen level during competitive interactions among stimuli.  
As a result, we  observe that given a specific stimulus, only a highly selective, small subset of neurons will activate \cite{shoham2006silent}.

\subsection{Background in Face Areas}
\textbf{Theme 2: Faces are processed in a unique pathway.}  The area responsible for face processing was first discovered by Sergent et al. \cite{sergent1992functional}.  This area was later named the fusiform face area and shown to activate more when people see faces rather than general objects \cite{kanwisher2010functional}.   Studies with patients suffering from prosopagnosia, the inability to recognize the faces of familiar people, have provided convincing evidence  that the recognition of faces and objects rely on distinct mechanisms \cite{dailey1999organization}.  




It is important to note that while faces activate specialized areas with the brain, these areas are not completely silent when non-face objects are viewed.  Instead, the cortical response for the preferred category is about \textit{twice that for the non-preferred category} as consistently observed in most normal individuals \cite{plaut2011complementary}.  The FFA can be activated by a range of non-face stimuli, such as houses and cars, and even novel objects introduced such as ``Greebles'' \cite{gauthier1999activation}.

\subsection{Background in Face and Visual Processing}
\textbf{Theme 3: Faces are processed in a holistic, coarse-to-fine manner.} 
In addition to the unique pathway for face processing, the holistic manner by which the face is recognized has also been discovered.  Maurer et al.  \cite{maurer2002many} notes that face stimuli are processed as a gestalt, and holistic processing occurs with the internal structure of the face and with the external contour.  Even simple circles containing three curved lines, if shaped like a smiling face, triggered a holistic face response.  The holistic response was thought to contribute to an early stage process so that one could distinguish faces from other competing objects \cite{taubert2011role}. 

The holistic approach of face recognition is also consistent with a related theory where research shows that the visual system integrates visual input in a coarse-to-fine manner (CtF) \cite{bar2006top}.  The CtF hypothesis states that low frequency information is processed quickly first, which then projects to high level visual areas.  Critically, the high level areas \textit{generate a feedback signal} that guides the processing of the high frequency (details) of the image \cite{petras2019coarse}.  

The observations that global information precedes finer details are consistent with the concept of inhibition between neurons having similar response properties.  These dynamics suggest a feedback or competitive process, whereby neurons that respond best to a given stimulus inhibit other neurons, resulting in a winner-take-all situation \cite{tsao2008mechanisms}.

\subsection{Background in Top-down Feedback}
\textbf{Theme 4: Top-down feedback is an essential component to visual perception.}
The neural circuitry within the brain imply that lateral and top-down feedback connections play a significant role in the visual pathway.  Feedback originating in higher-level areas such as V4 or IT can modify and shape V1 responses, accounting for contextual effects \cite{czigler2010unconscious}.  Feedback connections are thought to be critical to the function especially in order to enable inference about image  information from weak or noisy data \cite{dicarlo2012does}.  Additionally, there are competitive effects that can be seen through feedback connections.  Competition among objects is biased by bottom-up perception and top-down influences, such as selective attention \cite{kastner2001neural}.  In this case, feedback can not only enhance processing of the stimulus but also resolve competition between stimuli in the receptive field \cite{desimone1995neural}.


In the context of computer vision, there are only a few works that have addressed lateral and top-down feedback explicitly in a model.  In one case, top-down feedback was implemented using a parallel neural network to provide feedback to a standard CNN \cite{sam2018top}.  In another case, a novel feedback layer was introduced where the variables were activated by top-down messages, instead of bottom inputs \cite{cao2015look}.   While computationally expensive, top-down feedback could have significant impact on the performance of computer vision models \cite{kim2018deep}.  Elsayed et al. \cite{elsayed2018adversarial} showed that adversarial examples can fool time-limited humans, but not no-limit humans, stating no-limit humans are fundamentally more robust to adversarial examples and achieve this robustness via top-down or lateral connections.  And indeed, given both inhibitory and excitatory top-down feedback in a generative model,  immunity to adversarial examples was demonstrated \cite{kim2020modeling}.



\section{Methodology}
Given these overarching themes in neuroscience, e.g. selectivity through competition, dedicated pathways, holistic/coarse-to-fine processing, and top-down feedback, we were inspired to build and investigate a model that faithfully mimics these concepts.  It was requisite for our computational model to incorporate competition and feedback at multiple levels in its hierarchy.  In the following sections, we describe the algorithms and architectures that comprise our framework, and evaluate the performance of the model with respect to the experimental literature.  Lastly, we highlight and demonstrate the impact that our model can have on various computer vision applications.

\subsection{Sparse Coding for Selectivity and Competition}

The main algorithm that underlies our framework is sparse coding.   Sparse coding was first introduced by Olshausen and Field \cite{olshausen1997sparse}, in order to explain how the primary visual cortex efficiently encodes natural images.  
Sparse coding seeks a minimal set of generators that most accurately reconstruct each input image.  Each generator adds its associated feature vector to the reconstructed image with an amplitude equal to its activation.    For an input image, the optimal sparse representation is given by the vector of sparse activation coefficients that minimizes both image reconstruction error and the number of non-zero coefficients.   Since sparse coding minimizes reconstruction error, it can be considered as a form of self-supervised learning where classification labels are not necessary to train the system.  

Mathematically, sparse coding can be defined as follows.  Assume we have some input variable $x^{(n)}$ from which we are attempting to find a latent representation $a^{(n)}$ (we refer to as ``activations'') such that $a^{(n)}$ is sparse, e.g. contains many zeros, and can reconstruct the input, $x^{(n)}$, with high fidelity.  The sparse coding algorithm is defined as,
\begin{equation}
	\min_\Phi \sum^{\mathcal{N}}_{n=1} \min_{a^{(n)}} \frac{1}{2} \| x^{(n)} - \Phi a^{(n)}\|^2_2 + \lambda \|a^{(n)}\|_1
	\label{eq:sparsecode}
\end{equation}

Where $\Phi$ is the overcomplete dictionary, and $\Phi a^{(n)} = \hat{x}^{(n)}$, the reconstruction of $x^{(n)}$.  The $\lambda$ term controls the sparsity penalty, balancing the reconstruction versus sparsity term.  $\mathcal{N}$ is the total training set, where $n$ is one element of training.   $\Phi$ represents a dictionary composed of small kernels that share features across the input signal.   

There are a number of solvers for Equation \ref{eq:sparsecode}, but we selected the solver that is biologically informed.  This solver is the Locally Competitive Algorithm (LCA) \cite{rozell2007locally} that evolves the dynamical variables (neuron's membrane potential) when presented with some input stimulus.  Of particular importance is that the activations of neurons in this model compete and laterally inhibit units within the layer to prevent them from firing.  The input potential e.g. excitatory drive to the neuron state is proportional to how well the image matches the neuron's dictionary element, while the inhibitory strength is proportional to the similarity of the current neuron and competing neuron's convolutional patches, forcing the neurons to be decorrelated.  The LCA model is an energy based model similar to a Hopfield network \cite{hopfield1984neurons} where the neural dynamics can be represented by a nonlinear ordinary differential equation. 

We define the internal state of a particular neuron, $m$, as $u^m$ and the active coefficients as,
\begin{equation}
    a^m = T_\lambda(u^m)=H(u^m-\lambda)u^m
\end{equation}

where $T$ is an soft-threshold transfer function with threshold parameter, $\lambda$, and $H$ is the Heaviside (step) function.  

If we consider an input signal, $I$, The dynamics of each node is determined by the ordinary differential equation,
\begin{equation}
 \dot{u}^m =  \frac{1}{\tau} \bigg[ -u^m + (\Phi^T I) - (\Phi^T \Phi a - a^m)\bigg]
 \label{eq:orig}
\end{equation}
The $-u^m$ term is leaking the internal state, $\tau$ is the time constant, the $(\Phi^T I)$ term is ``charging up'' the the state by the inner product (match) between the dictionary element and input signal, and the $(\Phi^T \Phi a - a^m)$ term represents the competition from the set of active neurons proportional to the inner product between dictionary elements.  The $- a^m$ in this case is eliminating self interactions.  In summary, neurons that are selective to the image stimulus charge up faster, then pass a threshold of activation.  Once they pass the threshold, they begin to compete with other neurons to claim the representation.  Thus sparse coding with the LCA solver creates a sparse representation of selective neurons that compete with each other to represent stimuli \cite{paiton2020selectivity}. 

\subsection{Construction of Face and Object Pathways}
 \begin{figure}[th]
    \includegraphics[width=8.15cm]{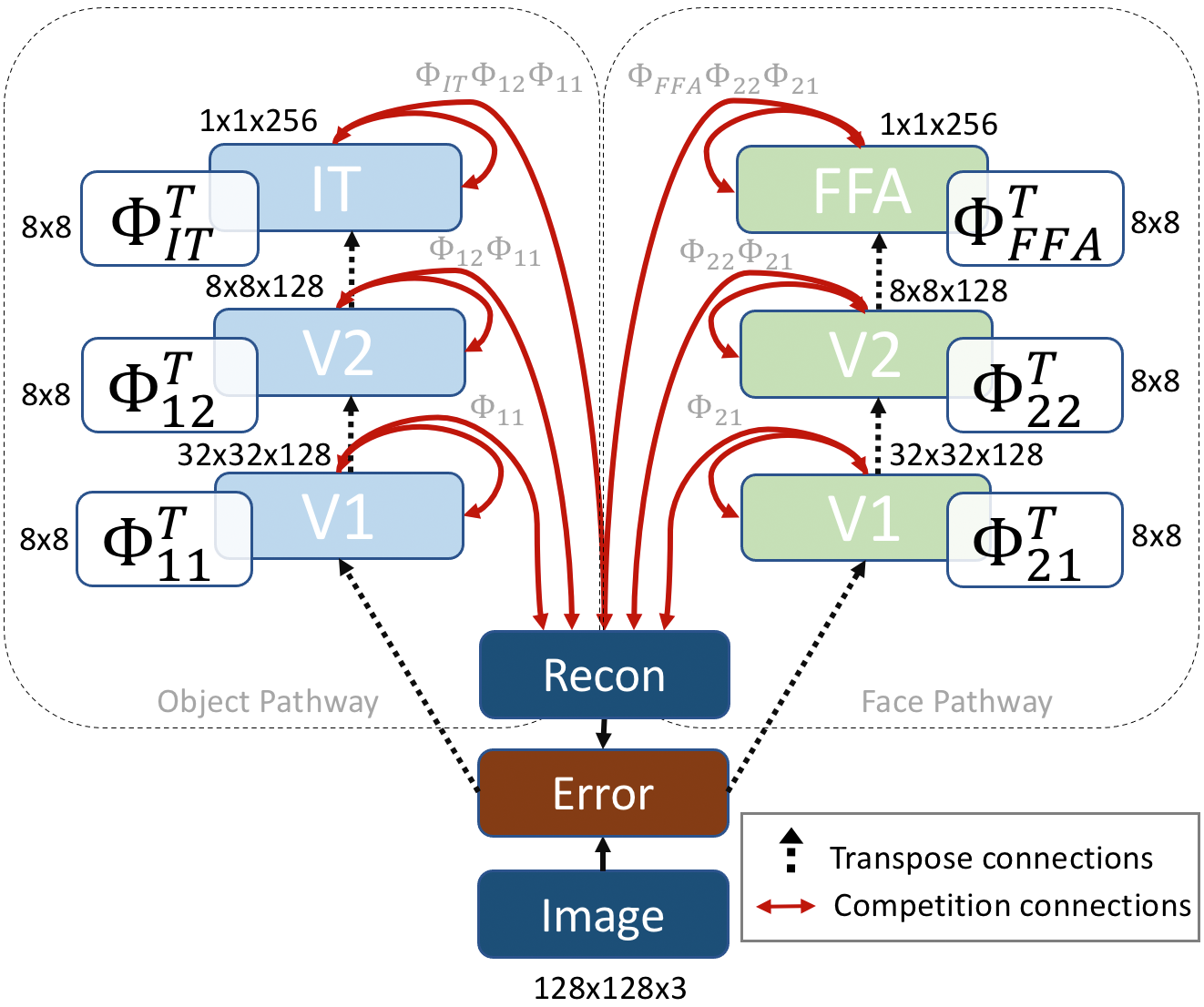}
  \caption{An illustration of our multipath deconvolutional competitive algorithm (MDCA) model.  Our model consists of two distinct pathways, one for faces and one for general objects.  Not only do the neurons in each of the layers compete to represent an input stimuli, every hierarchical layer in both pathways compete in the reconstruction of the input image. }
\label{fig:mdca}
\end{figure}

In the context of face recognition, there is strong suggestive evidence that some component of face processing is innate to the human visual system.  Psychologists have observed that newborn babies attend to faces, even within 1 hour of birth \cite{johnson1991biology,farroni2005newborns}.  Thus, for our model, we choose to pre-train each pathway to reflect the propensity of neural pathways we see in the visual cortex.  One pathway is tuned to reconstruct faces, and the other pathway tuned to reconstruct general objects.  The pathway consists of a 3-layer hierarchical, multiscale, convolutional sparse coding network as shown delineated by the dotted lines in Figure \ref{fig:mdca}.  The training procedure involved showing 10,000 images from the ImageNet \cite{deng2009imagenet} dataset and 10,000 images from the Celeb-A \cite{liu2015faceattributes} dataset to the respective pathways.  At the pre-training stage, the pathways are independent from each other and do not compete.

The images shown to the network have been resized to 128x128x3.  The dictionary sizes, activation maps, and architecture are identical in the two pathways.  There are 128, 8x8xC, (C being the number of input channels), dictionary elements in each of the respective V1 layers, $\Phi_{11}$ and $\Phi_{21}$ in Figure \ref{fig:mdca}. We stride by 4 throughout the hierarchy, thus increasing the receptive field of neurons by a factor of 4x at each layer. We keep the same size dictionary patches for V2, but at the top layer, FFA and IT, we expand the number of neurons to 256.  While this number is empirically chosen, it does have a biological connection as it has been shown that faces can be linearly reconstructed using responses of approximately 200 face cells \cite{chang2017code}.

\begin{figure}[b]
 \centering
     \subfigure[Face V1, $\Phi_{21}$ ]{\includegraphics[width=3.5cm]{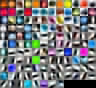}}\hspace{0.7cm}
    \subfigure[Object V1, $\Phi_{11}$]{\includegraphics[width=3.5cm]{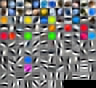}}
    \subfigure[Activity trigger average of the top (FFA) layer in the face pathway.]{\includegraphics[width=8.0cm]{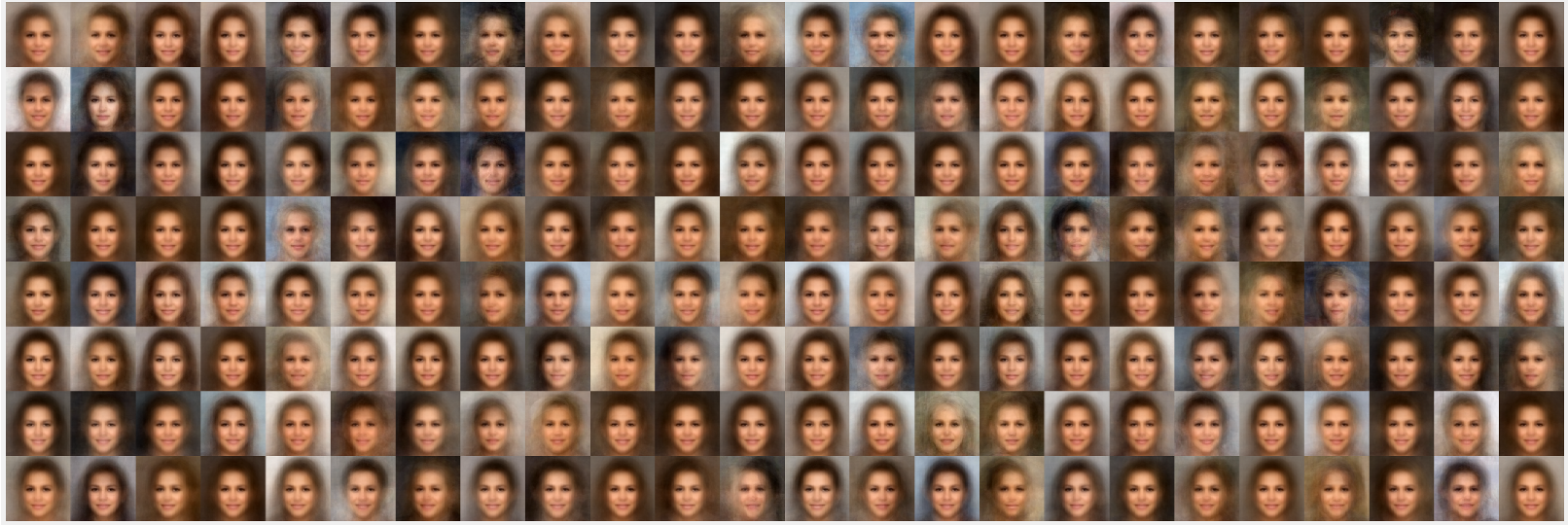}}
    \subfigure[Activity trigger average of the top (IT) layer in the object pathway.]{\includegraphics[width=8.0cm]{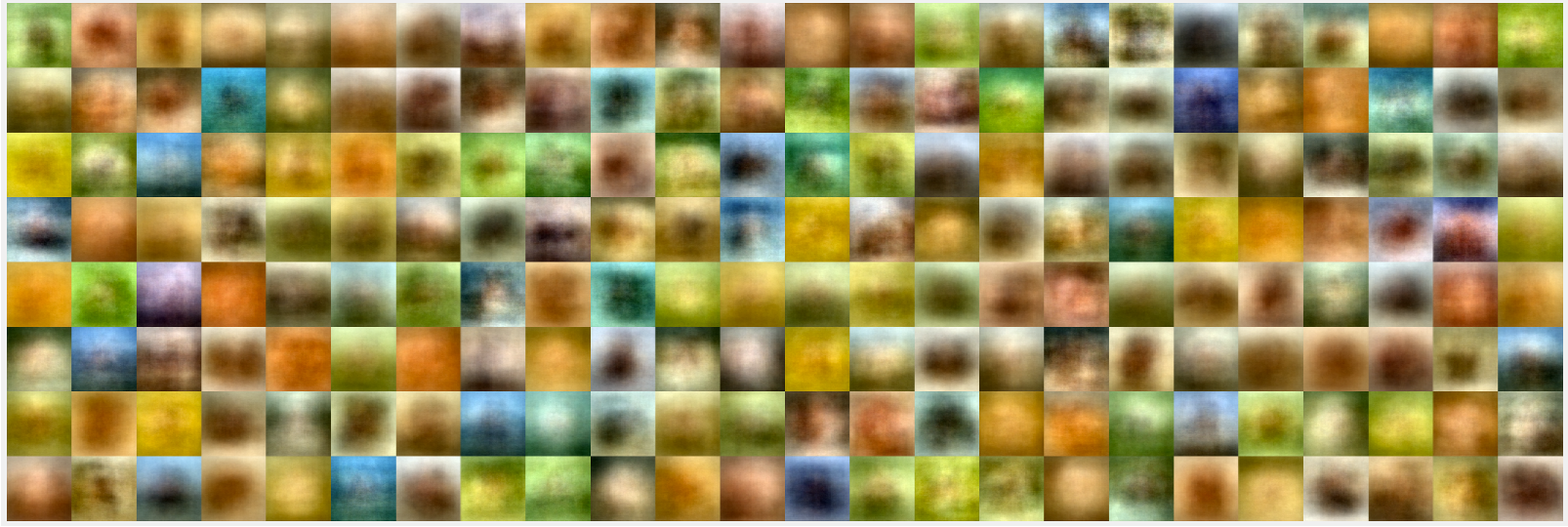}}
  \caption{Visualization of the learned dictionary elements at V1 and activity triggered averages of our top level (c) face and (d) object pathway.  The face pathway was shown images from Celeb-A and the object pathway was shown images from ImageNet.}

\label{fig:dictionary}
\end{figure}

Self-supervised learning of features can be obtained by taking the gradient of the reconstruction error with respect to $\Phi$, resulting in a biologically plausible local Hebbian learning rule.  The dictionary can be updated via Stochastic Gradient Descent (SGD).  At the low level, it is clear that both pathways learn similar features, edge detectors, color blobs, gradients, etc.  Examples of the dictionary learned at the V1 level can be seen in Figure \ref{fig:dictionary}(a)(b).    
\begin{figure*}
    \centering
    \includegraphics[width=16.15cm]{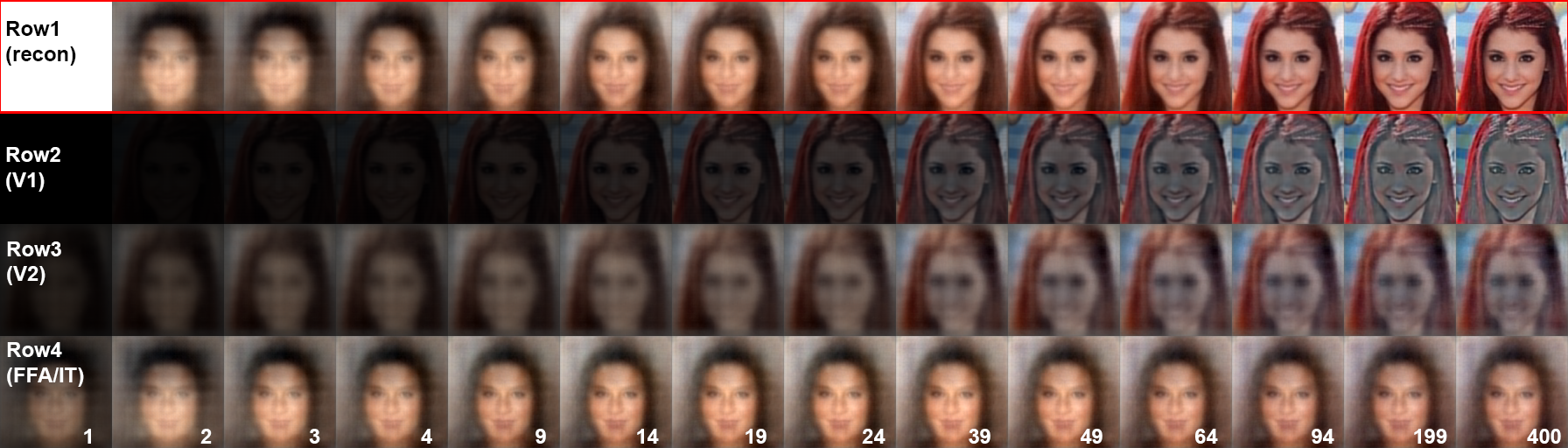}
    \caption{Illustration of the coarse-to-fine reconstruction over 400 timesteps.  \textbf{(Row 1)} shows the total reconstruction at the given timestep.  \textbf{(Row 2)} is the summed contribution from V1 in both pathways. \textbf{(Row 3)} is the summed contribution V2 in both pathways, and \textbf{(Row 4)} is the summed contribution of FFA and IT.  The numbers in the bottom right corner indicate the timestep of the each column.  The holistic, coarse-to-fine approach can be easily seen by the contributions at different hierarchical levels.}
    \label{fig:progression}
    \vspace{-0.3cm}
\end{figure*}

To understand the higher levels of the network, we compute an activity triggered average (ATA) akin to spike-triggered averages that have been used to characterize the response properties of a neuron to a time-varying stimulus.  Thus, the ATA can be thought of as the average response of a neuron when shown a set of input images.  We can see at the top level of the network, these neurons are specialized to global, holistic information of faces or objects, see Figure \ref{fig:dictionary}(c)(d).  In the FFA region, we see a heterogeneity of orientation, background, and hair but an overall homogeneity in face type.  For the IT region, we see color differences, scene-like backgrounds, and ghostly foreground blobs due to the fact it was tasked to encode a dataset with higher variance given the same number of neurons as the FFA.

\subsection{Multipath Deconvolutional Competitive Algorithm (MDCA)}
While it is now fairly trivial and commonplace to build hierarchical structures in artificial neural networks, any element of top-down feedback is virtually non-existent, especially in the inference process.  However, feedback is ubiquitous in biological neural networks, and a core component in our model.  We combine paths together in a multiscale, hierarchical competitive structure that we refer to as the Multipath Deconvoloutional Competitive Algorithm (MDCA). The learning of elements at multiple scales was explored by \cite{mairal2007multiscale} and used for image and video restoration  \cite{mairal2008learning}.  Multi-scale dictionaries have been used in applications where it is important to capture the scale of certain features as in palmprint identification \cite{zuo2010multiscale}, audio feature learning \cite{dieleman2013multiscale}, and removing rain streaks at different scales \cite{li2018video}.  Our network consists of a deep deconvolutional sparse coding network similar to the work of Zeiler \cite{zeiler2011adaptive} and Paiton \cite{paiton2016deconvolutional}.  


At a conceptual level, our model is implementing the following process illustrated in Figure \ref{fig:mdca}.  An input stimulus is presented, and very quickly sent up the hierarchy of both pathways.  One could think of this initial step as a feed forward pass in a typical deep learning model.  Each neuron in every layer is ``charged up'' by the input stimulus, where neurons at higher levels have larger receptive fields, and neurons at the top level see the entire input stimulus.  As each neuron passes threshold, they add their respective feature to the reconstruction via deconvolution.  Thus, the reconstruction layer is not only influenced by fine, high-frequency features from the lower layer, but also guided by the large, low spatial-frequency activated features of the higher layer.  As the stimulus is reconstructed over time, the network computes the error between the input and reconstruction at each timestep.  This error is then forwarded up the hierarchy, driving the neurons to compete for the remaining residual representation.  In our experiments, we evolve our recurrent network over t=400 timesteps.

Mathematically, we define the reconstruction, $\hat{x}$, in our Multipath Deconvolutional Competitive Algorithm as the following,
\begin{equation}
	\hat{x}^{(n)} = \sum_{m=1}^{M}(\sum_{k=1}^{K}(\prod_{l=1}^{k} \Phi_{m,l}) a_{m,k}^{(n)})
	\label{eq:sparsecode}
\end{equation}
where $m \in M$ is the number of paths, and $l,k \in K$ is the number of multiscale layers in the neural network.  In our case shown in Figure \ref{fig:mdca}, we have $M=2$ e.g. object pathway and face pathway, and three multiscale layers, $K=3$.  Our reconstruction term for the first pathway is, $\Phi_{11}a_{11} + \Phi_{12}\Phi_{11}a_{12} + \Phi_{13}\Phi_{12}\Phi_{11}a_{13}$, and similarly, $\Phi_{21}a_{21} + \Phi_{22}\Phi_{21}a_{22} + \Phi_{23}\Phi_{22}\Phi_{21}a_{23}$, for the second path.

\section{Experiments and Results}
\subsection{Coarse-to-fine Information Flow}

 \begin{figure}[ht]
 \centering
    \subfigure[Magnitude Response]{\includegraphics[width=4.1cm]{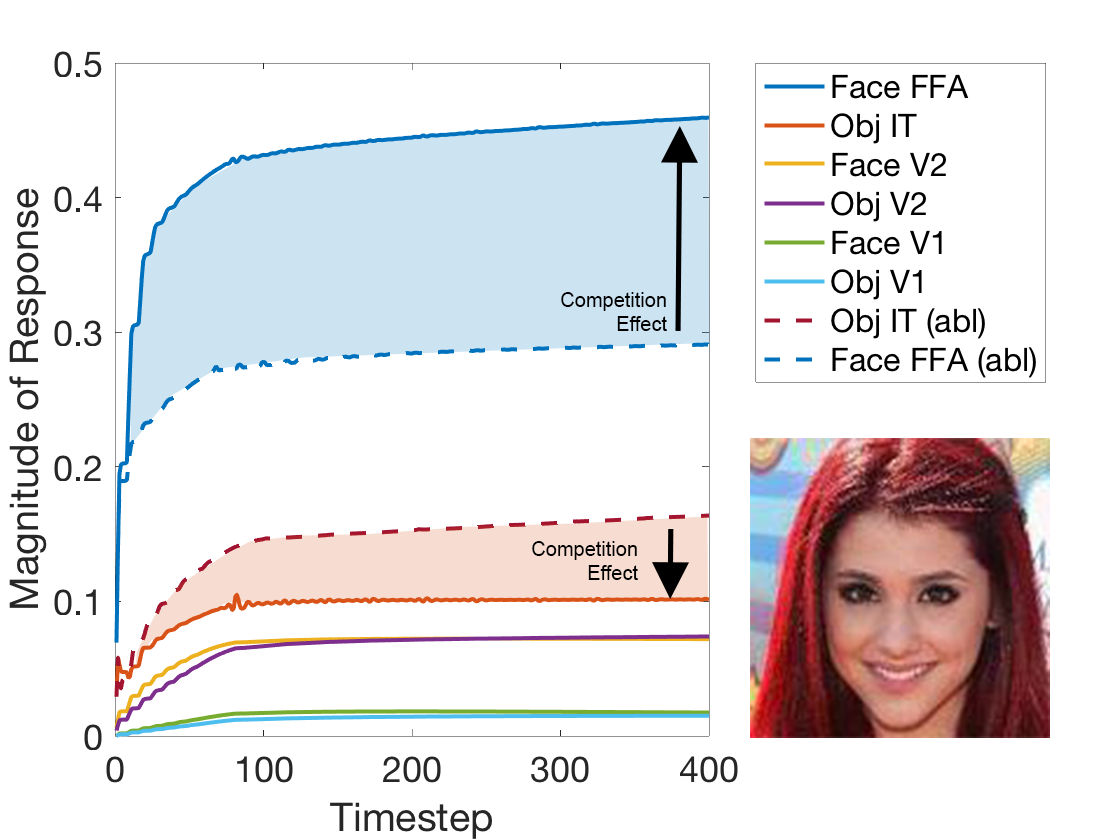}}
    \subfigure[Neuron Activity]{\includegraphics[width=4.1cm]{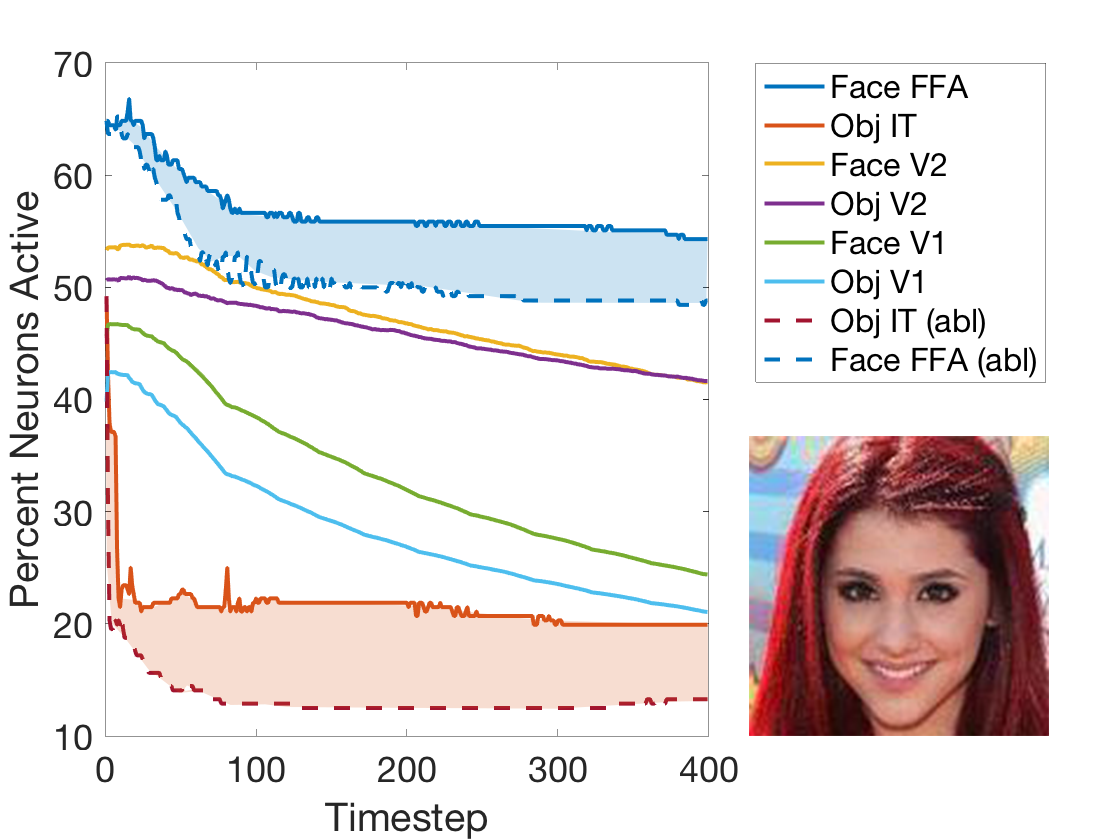}}
    \subfigure[Magnitude Response]{\includegraphics[width=4.1cm]{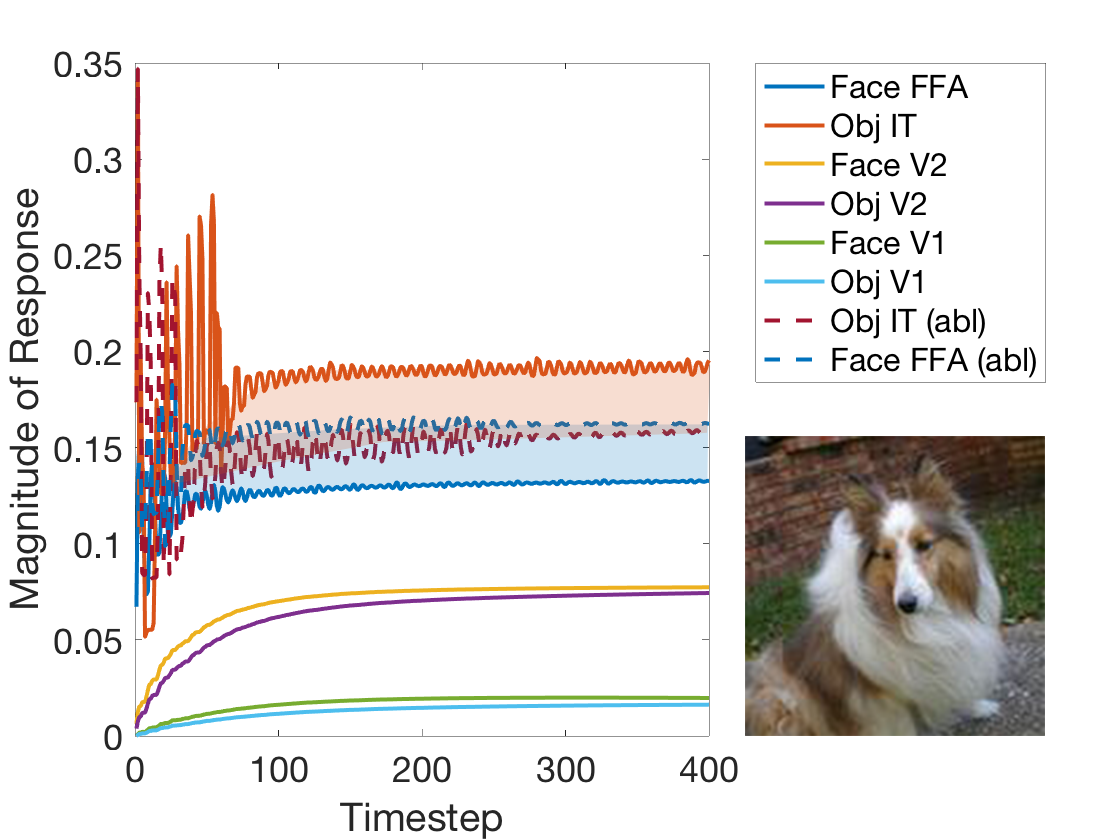}}
    \subfigure[Neuron Activity]{\includegraphics[width=4.1cm]{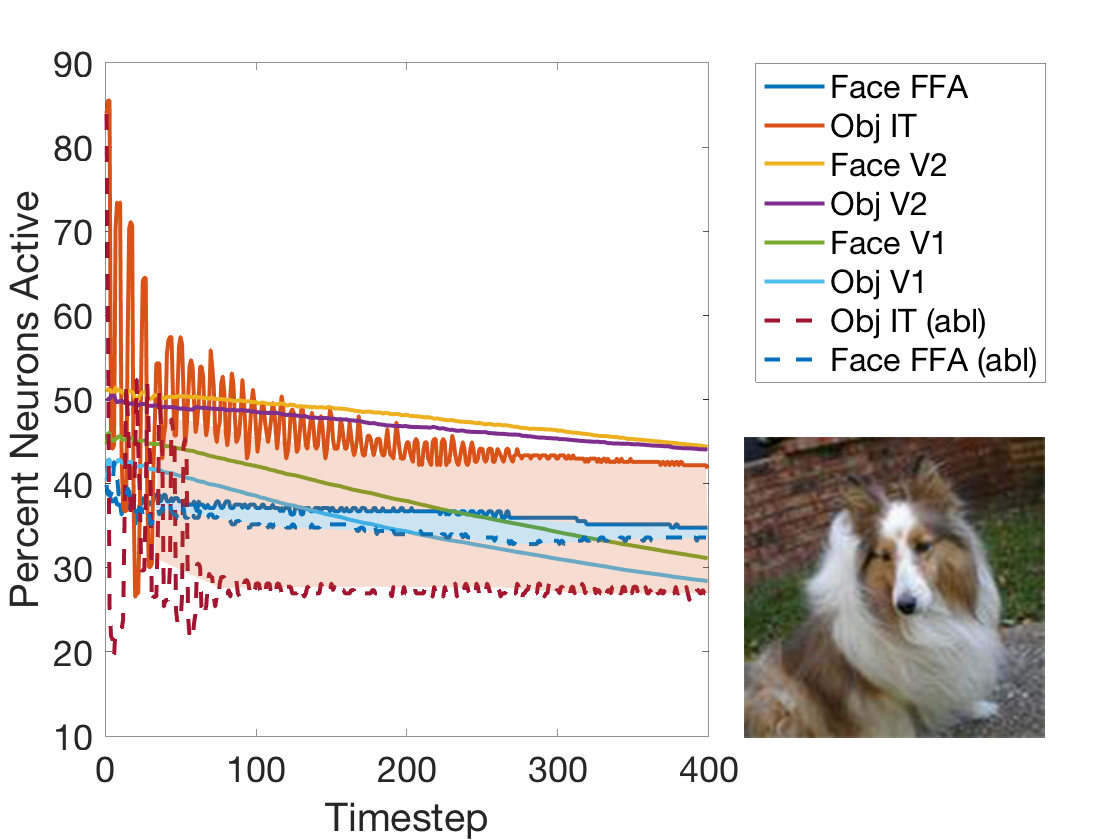}}
  \caption{We plot the (a)\&(c) magnitude of response and (b)\&(d) percent of active neurons in all of the MDCA layers over 400 timesteps.  The magnitude of response at the V1 and V2 levels are similar; however, there is over a 4x response in the Face FFA region compared to the Object IT region in (a).  Further, we see a the effect that competition has on the response (highlighted in blue and red), when compared to an ablation pathway that does not compete. }

\label{fig:responsecurves}
\end{figure}
In our first experiment, we investigate the activity of the MDCA network as it processes input stimuli.  Given an input image, the objective of the network is simply to minimize reconstruction error.  We can visualize the process of reconstruction in Figure \ref{fig:progression}, and quantify the response at all levels in all pathways of our network (Figure \ref{fig:responsecurves}).  In the reconstruction, we can see that our \textbf{MDCA network is reconstructing the image in a holistic, coarse-to-fine manner}, consistent with the CtF neuroscience literature \cite{bar2006top}.  We can clearly visualize that low frequency information is processed quickly first and high level areas \textit{generate a competitive feedback signal} that guides the processing of the high frequency (details) of the image \cite{petras2019coarse}. 

\subsection{Ablation of Pathway Competition}
We perform an ablation study to confirm that the selectivity of the neurons in our network is mainly due to the competition of multiple pathways.  In Figure \ref{fig:responsecurves}(a) we see the magnitude of response of the FFA increases nearly 2x, and suppresses the IT response on its preferred stimulus.  We see a similar pattern with Object IT when presented its preferred stimulus, Figure \ref{fig:responsecurves}(c). From these results, we gain insight on how the visual cortex is able to select and respond to a specific stimulus.  This can also explain why traceable and identifiable activity still occurs in areas of the brain that are non-selective for a particular stimulus \cite{mcgugin2012high}.  

\begin{figure}[ht]
 \centering
    \subfigure[1k ImageNet images]{\includegraphics[width=3.65cm]{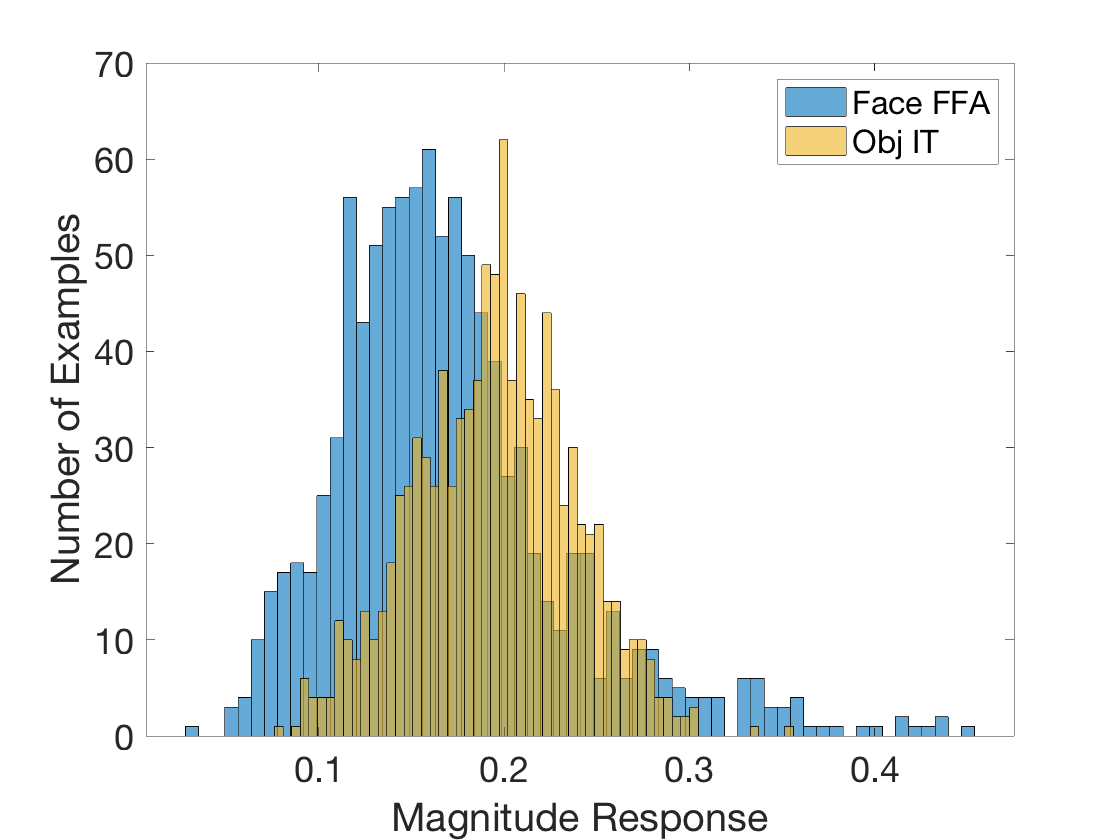}}
    \subfigure[1k Celeb-A images]{\includegraphics[width=3.65cm]{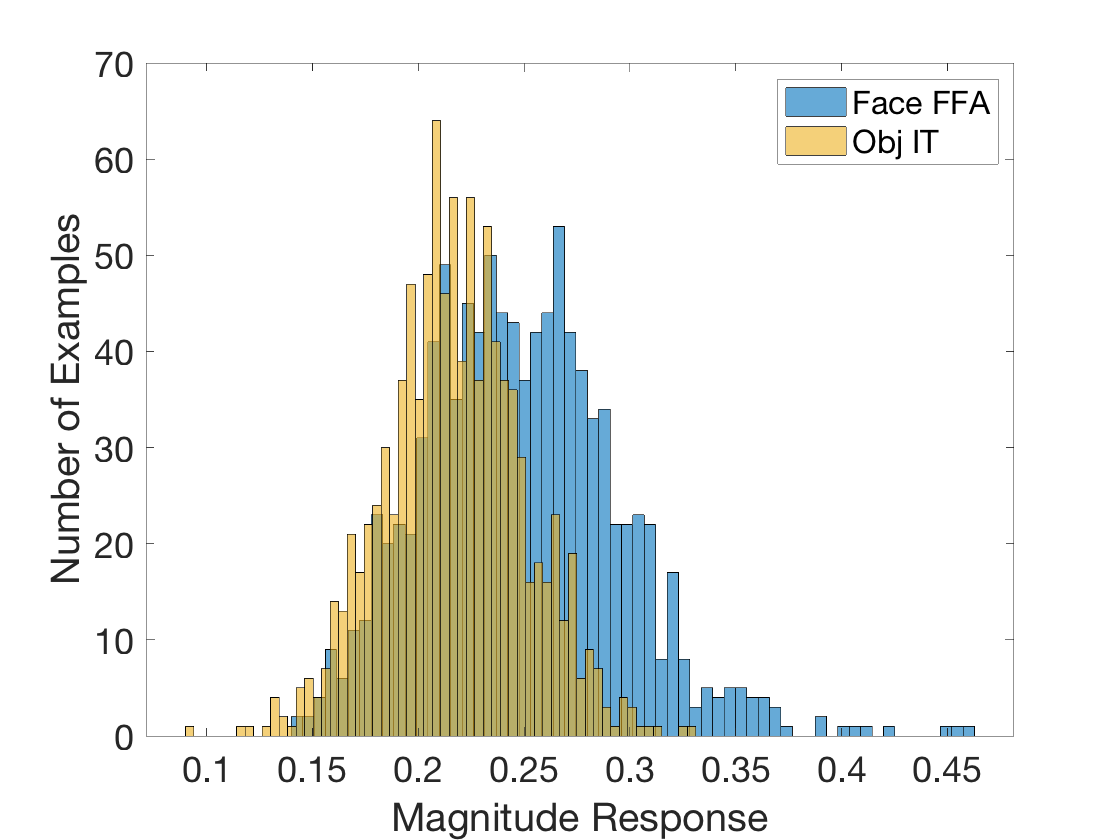}}

  \caption{Shows the independent magnitude of response from each pathway to (a) 1k ImageNet images, and (b) 1k Celeb-A faces.  Response overlap indicates minimal selectivity to specific stimuli.}
  \vspace{-0.4cm}
\label{fig:selective2}
\end{figure}

Next, we generalize the ablation study using 2,000 images (1k ImagNet, 1k Celeb-A).  We plot the magnitude of response from the two independent pathways in Figure \ref{fig:selective2}.  From these results, we can confidently say that \textbf{there is no significant selectivity in independent pathways.}  The pathways only become highly selective when they are required to compete for the input stimulus.

\subsection{Inverted Face Responses}
 \begin{figure}[ht]
 \centering
  \includegraphics[width=7.0cm]{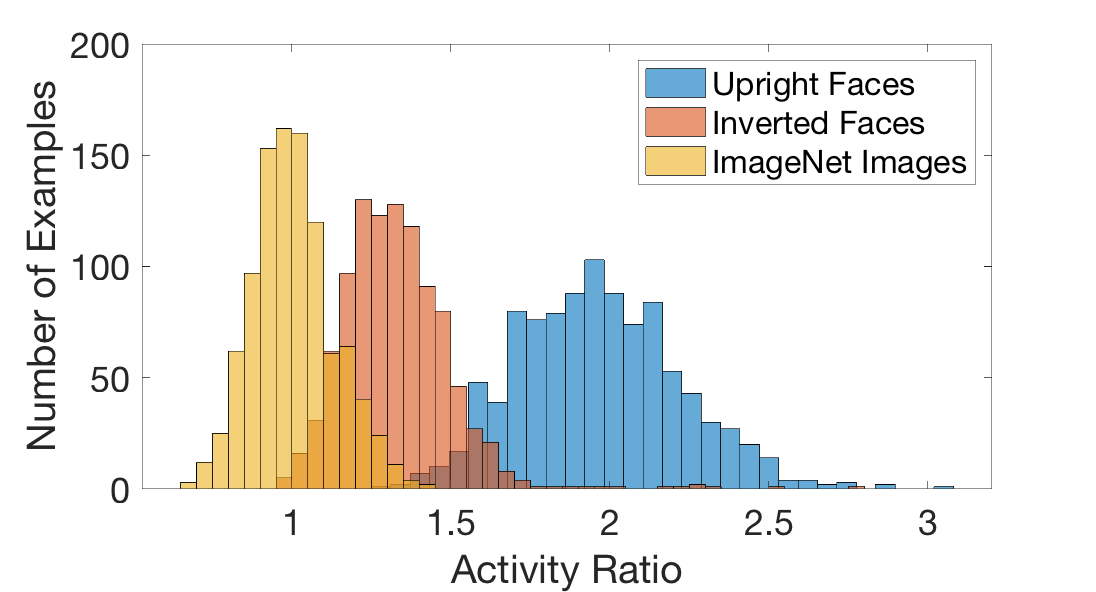}
  \caption{Histogram of the ratio of activity shown by the FFA and IT neurons for upright faces from Celeb-A, inverted faces e.g. upside down, and ImageNet images. The faces and objects can be easily separated by an activity threshold around 1.4.}

\label{fig:histogram}
\end{figure}
 \begin{figure*}[ht]
     \subfigure[Guide]{\includegraphics[width=4.6cm]{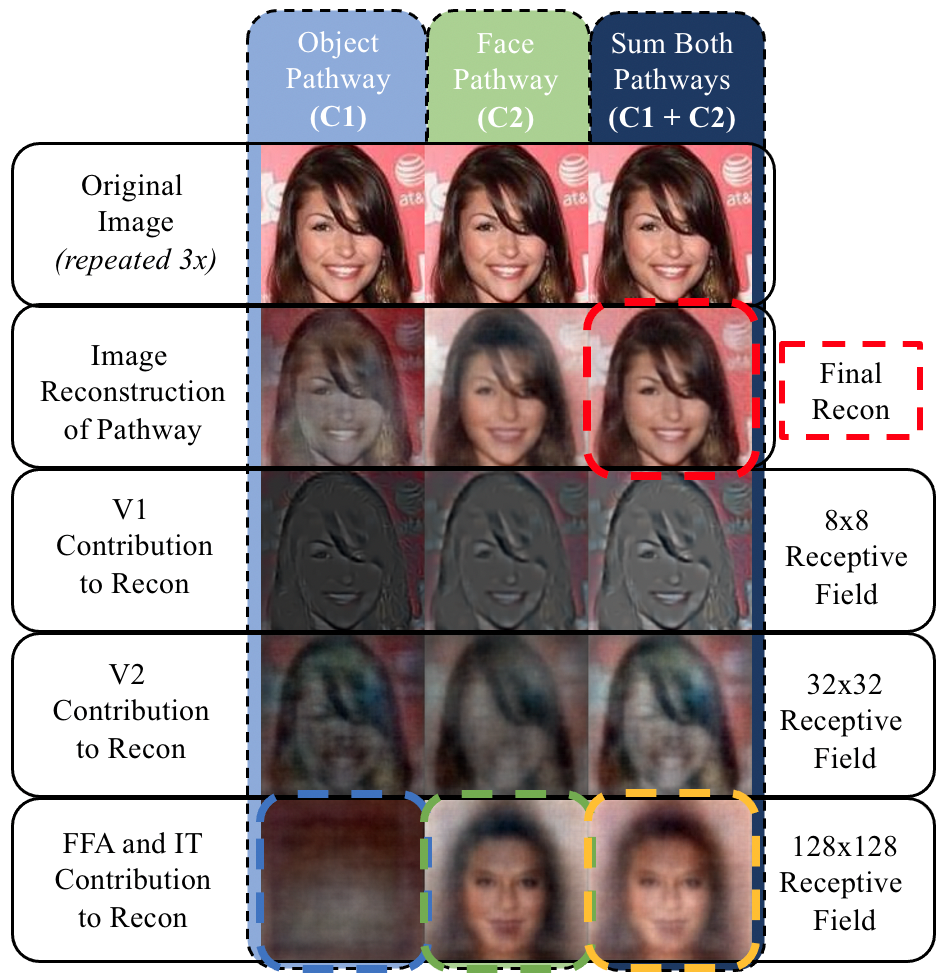}}
    \subfigure[Object Image]{\includegraphics[width=2.45cm]{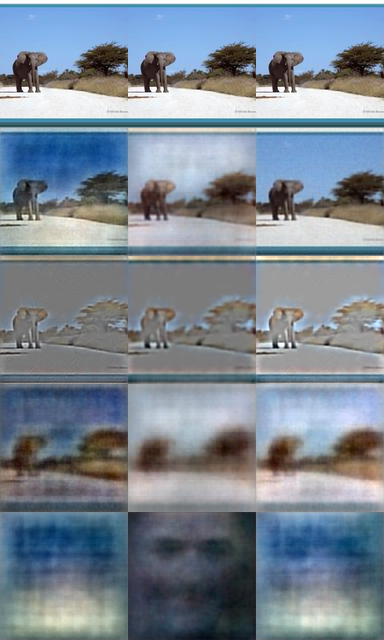}}
    \subfigure[Inverted Face]{\includegraphics[width=2.45cm]{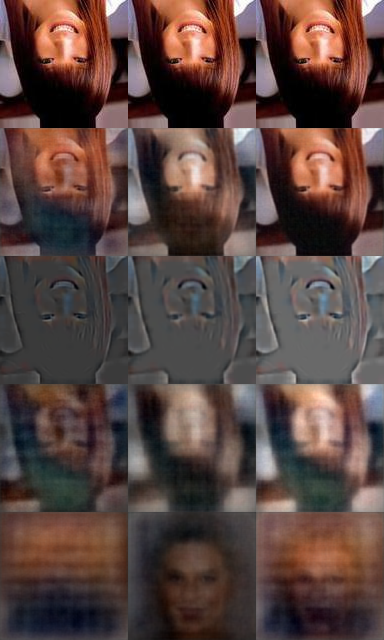}}
    \subfigure[Facephenes]{\includegraphics[width=2.450cm]{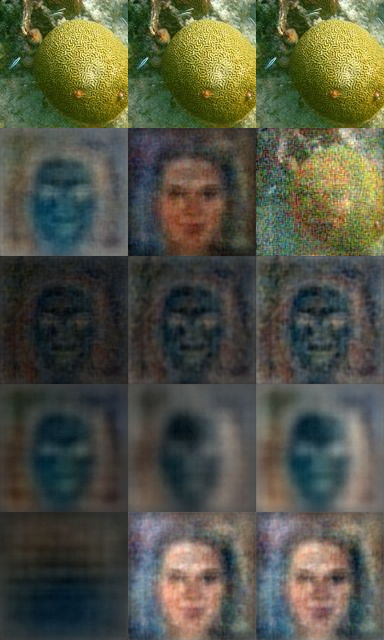}}
    \subfigure[Bias Ex.]{\includegraphics[width=2.450cm]{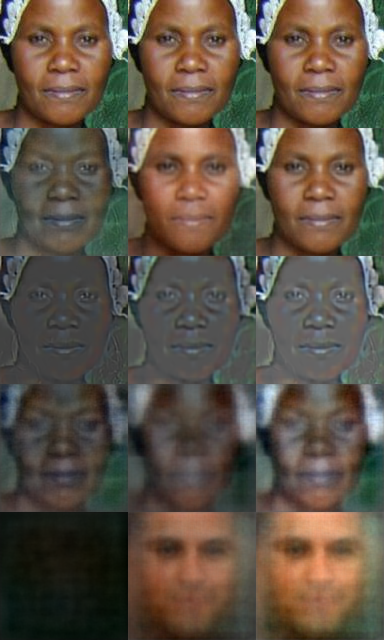} \label{fig:bias_res1} }
    \subfigure[Bias Ex.]{\includegraphics[width=2.450cm]{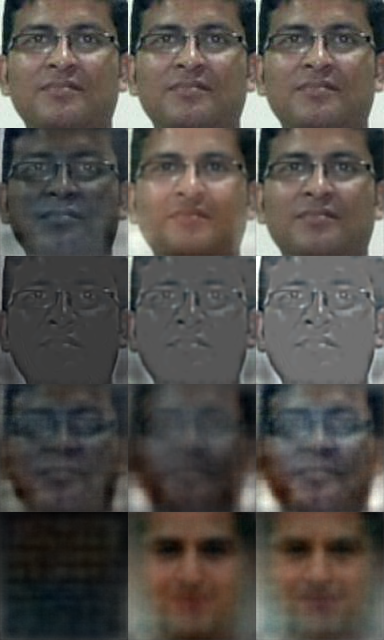} \label{fig:bias_res2} }

  \caption{(a) Guide to understanding the resulting reconstructions.  Specific images of interest are called out in dotted boxes e.g. (red) the final reconstruction, (blue) IT, top level response from the object pathway, (green) FFA, top level response from the face pathway, (yellow) IT+FFA, combined top level response from both pathways.  We show the reconstruction result of (b) an image from ImageNet.  In (c) we show the response of an inverted face.  Note that the FFA attempts to use an upright face, and the other layers are forced to adapt.  In (d) we show the effect of strongly (2000x normal stimulus strength) exciting the N-3 FFA neuron.  In (e) and (f) we show how our model still has bias and is forced to use white male representations in the FFA, yet can overcome it in the reconstruction. }

\label{fig:exp_results}
\end{figure*}

Following the ablation study, we are able to confirm that the addition of competitive pathways results in a roughly 2x magnitude of response of the FFA compared to the IT layer for the preferred stimulus at the top of both pathways, see Figure \ref{fig:histogram}.   This mirrors the fact that the cortical response for the preferred category is about \textbf{twice that for the non-preferred category} as consistently observed in most normal individuals \cite{plaut2011complementary}.  The histograms represent the response ratio for the same 1k ImagNet and 1k Celeb-A images.  The separation of distributions is clearly visible for the faces and non-face stimuli.  A simple maximum likelihood estimator on the ratio of activity can give us over 99\% classification accuracy on the binary task of face/not-face.

While upright faces evoke strong responses in the visual system, many researchers have found that there is a face inversion effect, where inverted e.g. upside down faces significantly delays the perceptual encoding of faces \cite{yarmey1971recognition}.  In the case of a brain damaged patient, they had no difficulty recognizing faces, but were unable to recognize faces that were inverted, supporting the idea that the FFA is specialized for processing faces in a normal orientation \cite{moscovitch1997special}.
Recently, fMRI and scalp ERP studies found inverted faces, ``slightly but significantly decreases the response of face-selective brain regions, including the so-called fusiform face area (FFA)'' \cite{rossion2002does}.  As we can see in Figure \ref{fig:histogram}, we are seeing a similar \textbf{slight but significant decrease in the response of the FFA to inverted faces.}  The reconstruction of an inverted face can be seen in Figure \ref{fig:exp_results}(c), where the FFA struggles to match an upright face to the stimulus.  Thus, the object IT layer is able to capture more of the representation.



\subsection{Facephenes - Patient Hallucinates Faces}

In this experiment, we revisit the initial neuroscience study on Facephenes.  Recall that this patient was being treated for intractable epilepsy and had electrodes placed on the fusiform face area.  While visually presented with objects, e.g. a ball, a book, a written character, the neurons within the FFA were artificially stimulated.  Remarkably, the perceived objects morphed, specifically, the patient saw parts of a face emerge on top of the inanimate objects. 

To replicate this experiment, we artificially stimulated a random neuron in the FFA (2000x the normal stimulus weight), while showing the network images from ImageNet and asking it to reconstruct what it sees.  We can see in Figure \ref{fig:exp_results}(d), that a face response is strongly elicited, and the other pathway and layers attempt to recover from the improper top-down feedback.  While the network does a respectable job in the final reconstruction, one can see face parts superimposed on top of the object.  As mentioned in the 2017 facephenes study, the patient reported no change in the object viewed, apart from the facephenes and rainbows apparently superimposed on them \cite{schalk2017facephenes}.  Additional neuronal response and stimulus weights can be seen in our supplemental material.  Thus, we demonstrate that our model \textbf{ consistent with the abnormal effects that can occur given an artificial or improper top-down feedback.}

\subsection{Robust Category Level Classification}
 \begin{figure*}[ht]
 \centering
    \subfigure[Black Male]{\includegraphics[width=2.8cm]{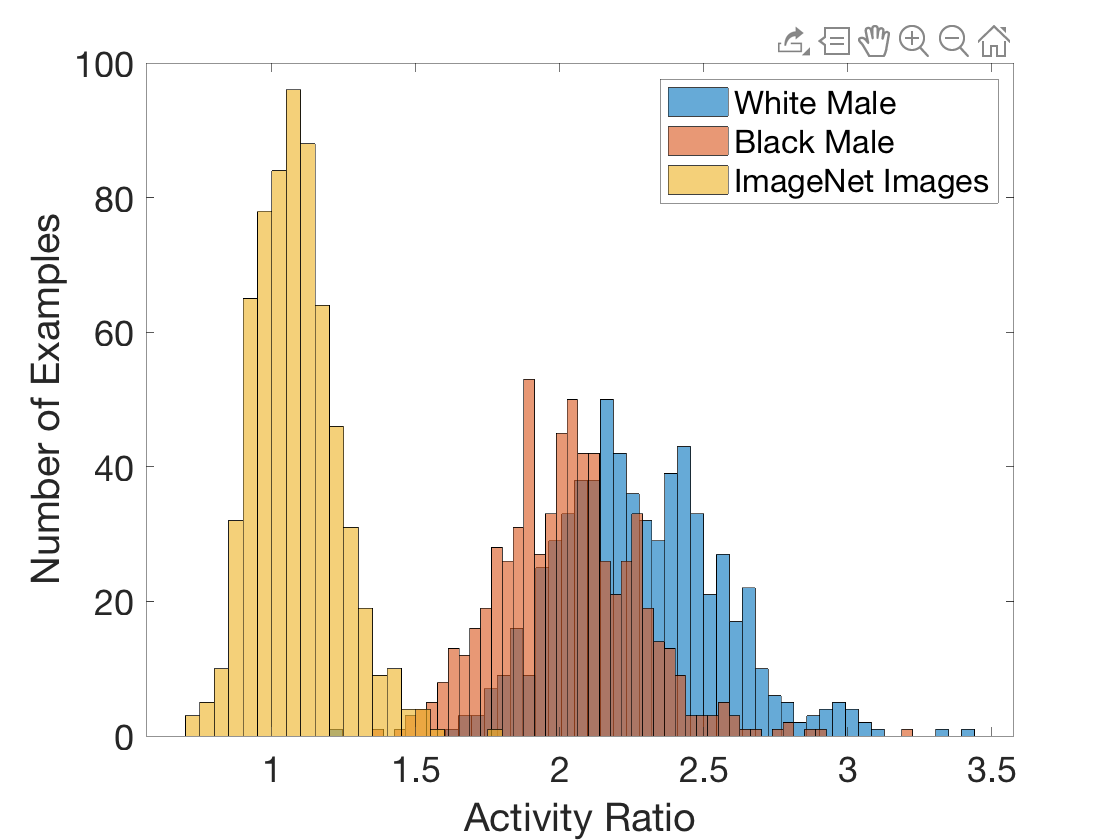}}
    \subfigure[Black Female]{\includegraphics[width=2.8cm]{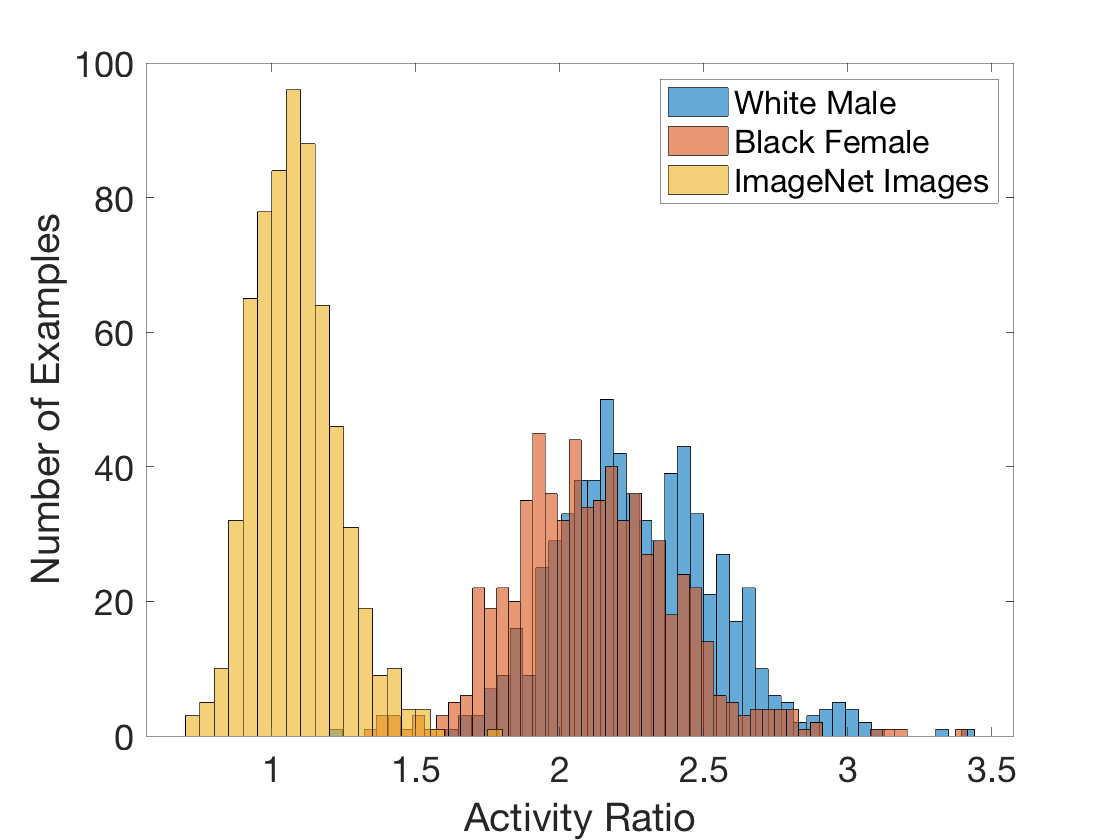}}
    \subfigure[East Asian Male]{\includegraphics[width=2.8cm]{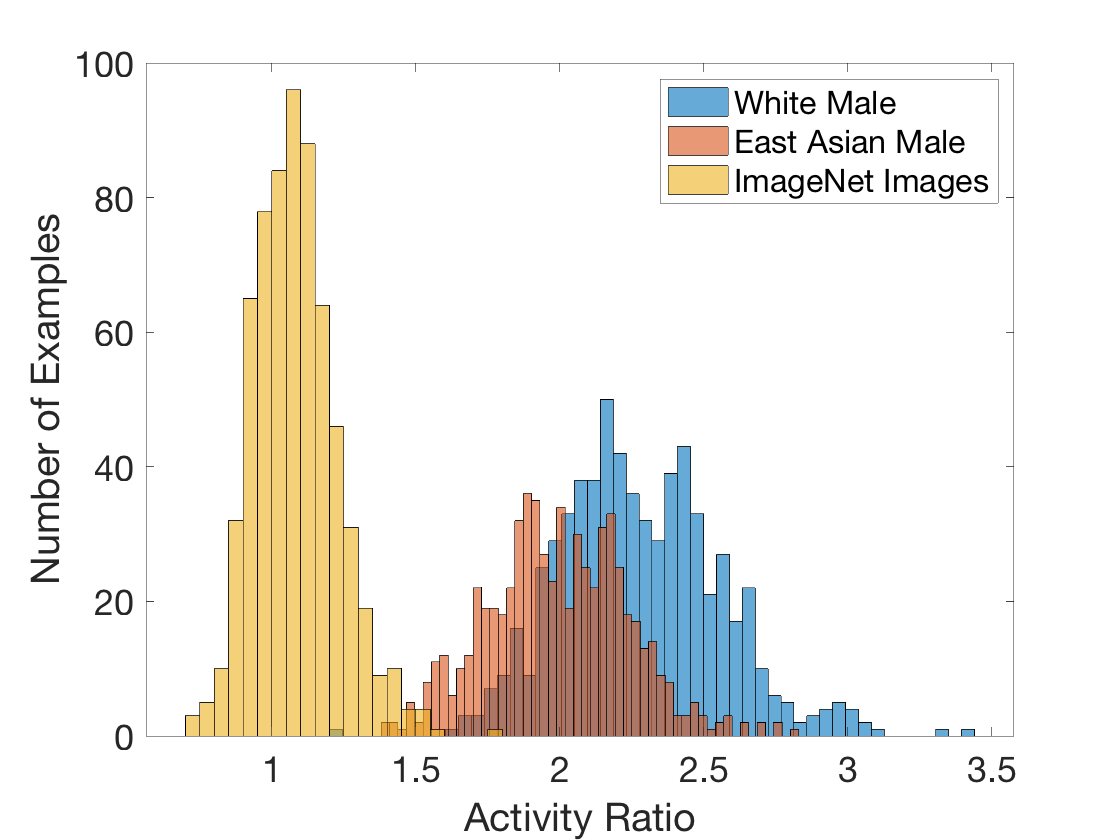}}
    \subfigure[East Asian Female]{\includegraphics[width=2.8cm]{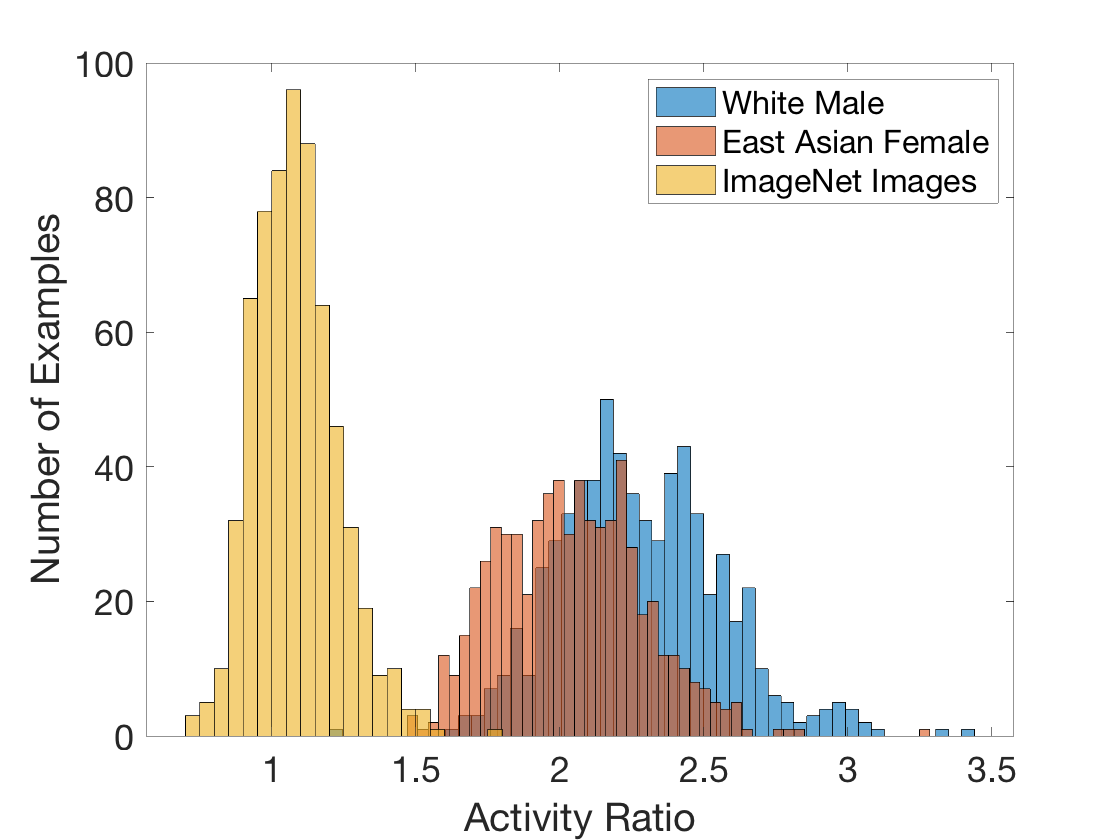}}
    \subfigure[Indian Male]{\includegraphics[width=2.8cm]{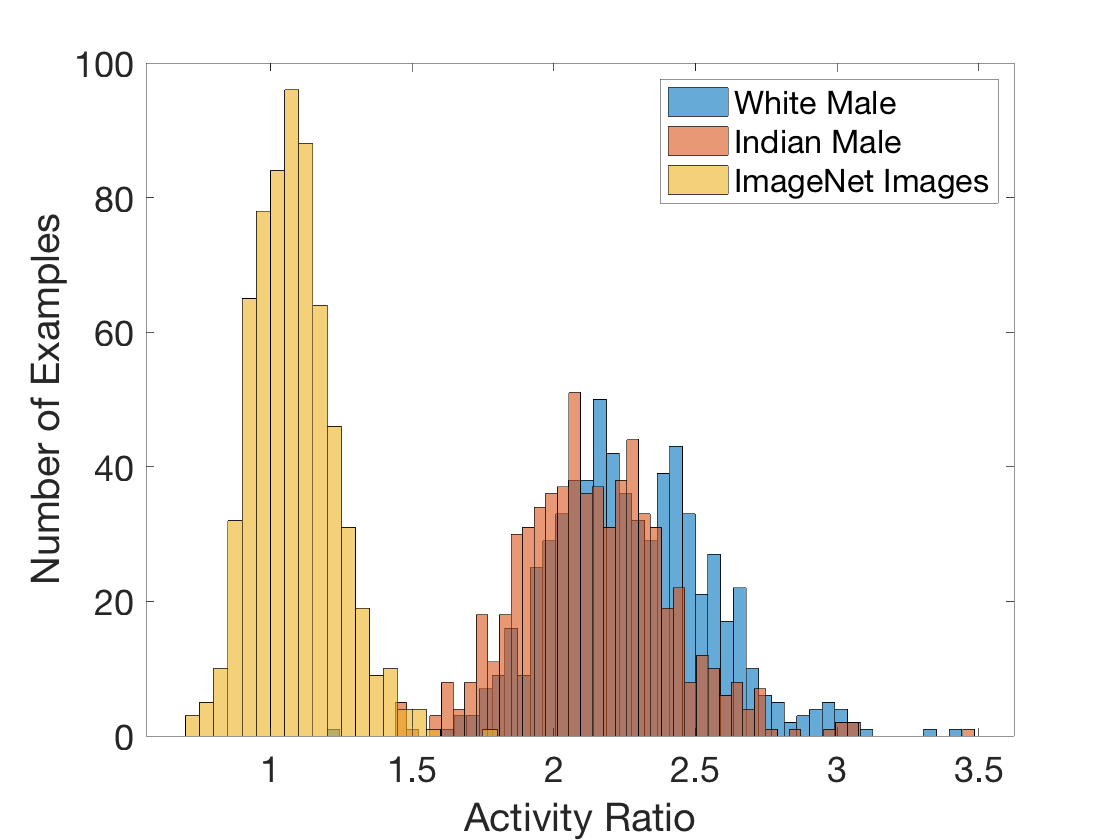}}
    \subfigure[Indian Female]{\includegraphics[width=2.8cm]{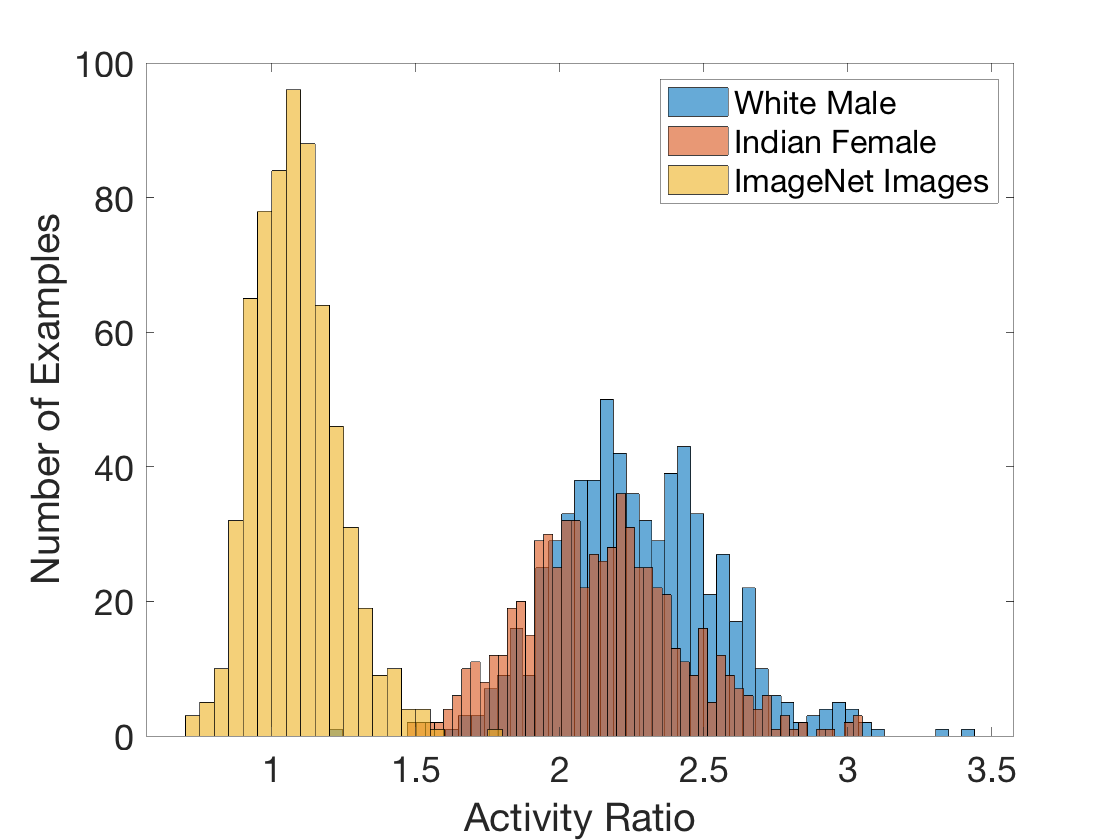}}
    \subfigure[Latino Male]{\includegraphics[width=2.8cm]{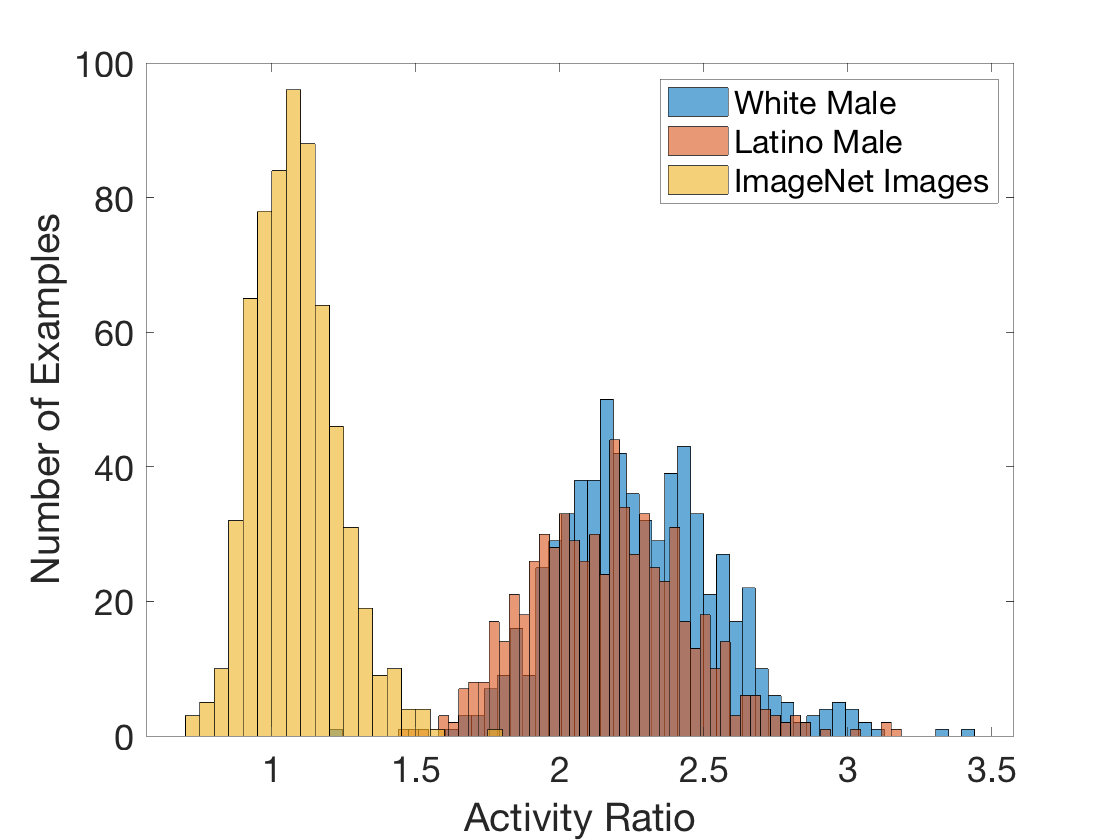}}
    \subfigure[Latino Female]{\includegraphics[width=2.8cm]{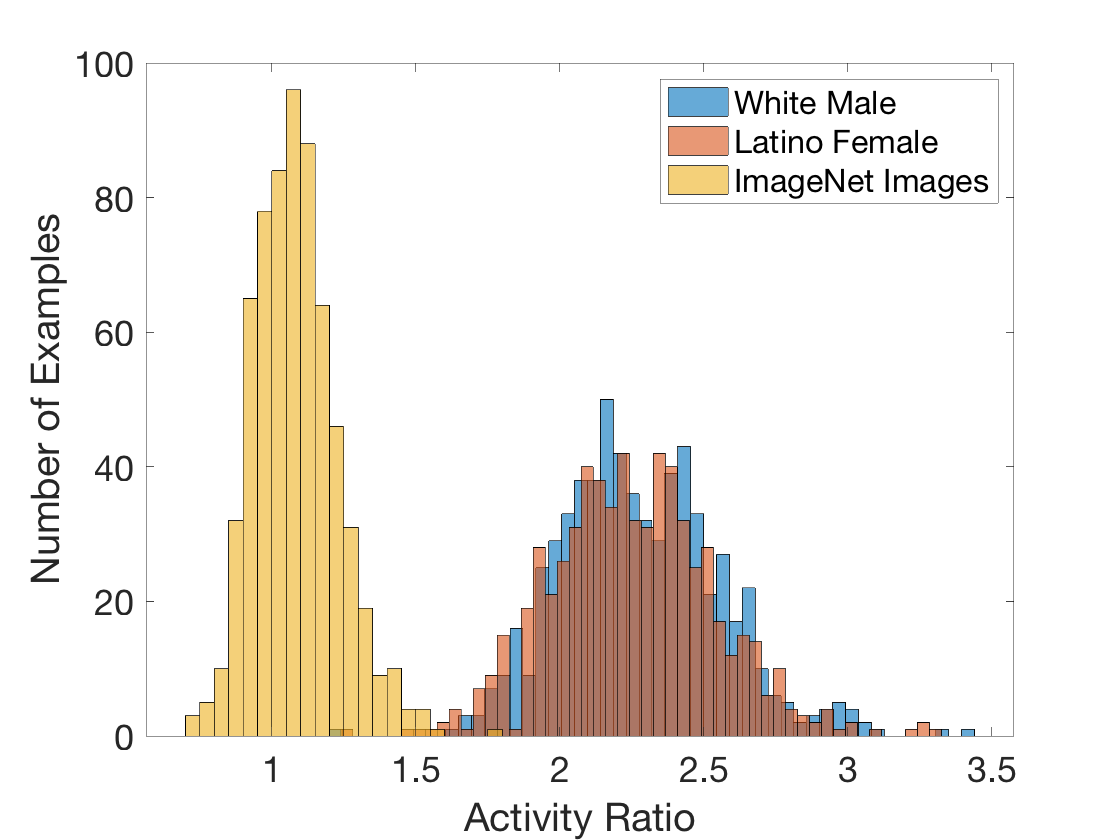}}
    \subfigure[Middle Eastern Male]{\includegraphics[width=2.8cm]{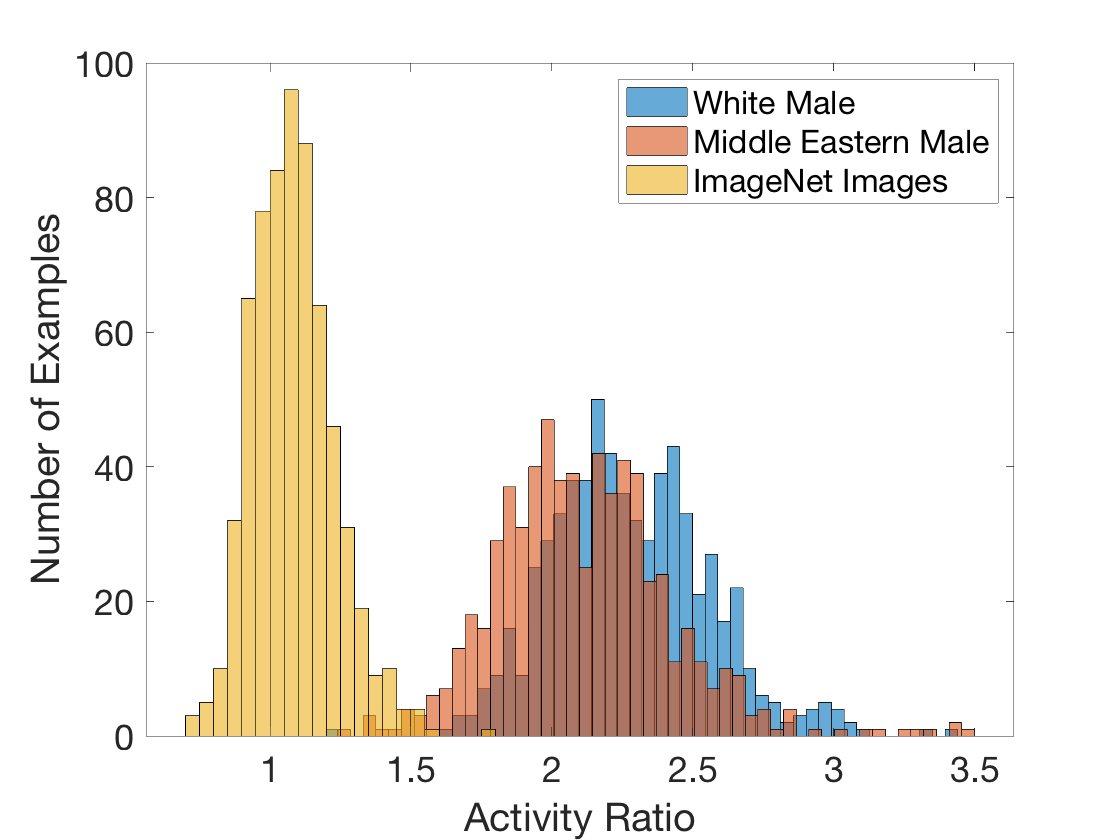}}
    \subfigure[Middle Eastern / F]{\includegraphics[width=2.8cm]{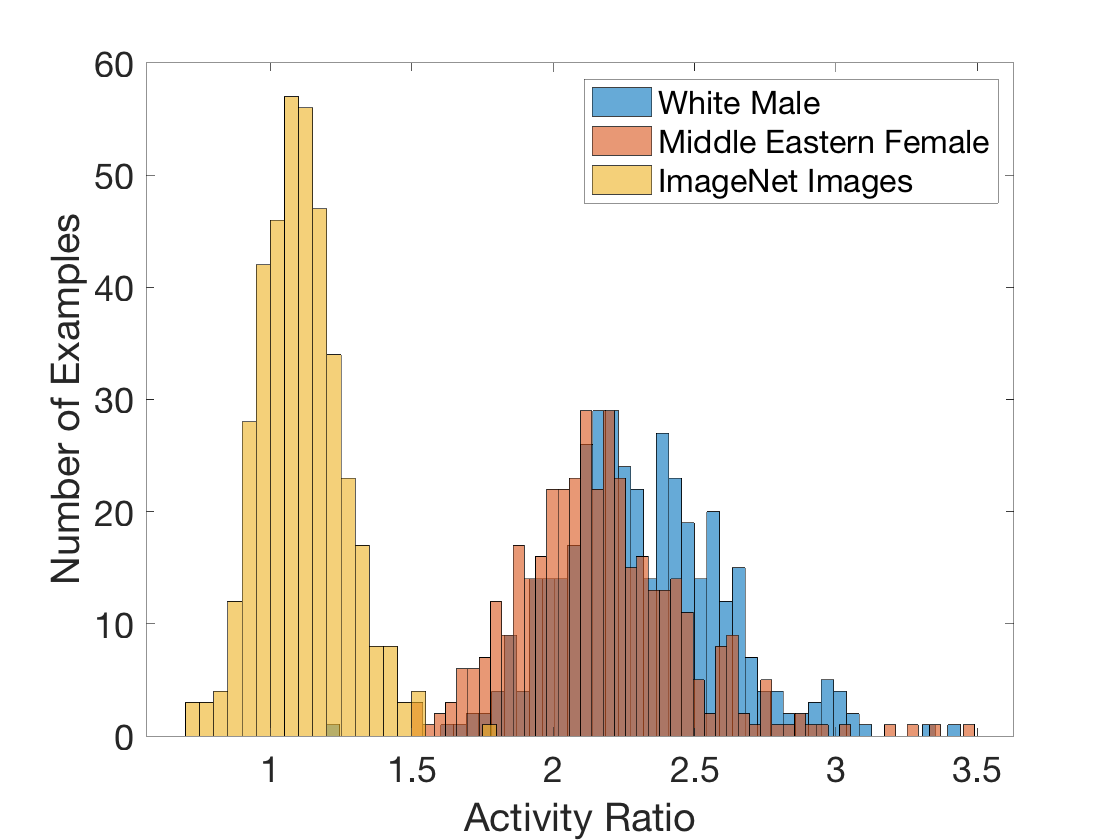}}
    \subfigure[Southeast Asian / M]{\includegraphics[width=2.8cm]{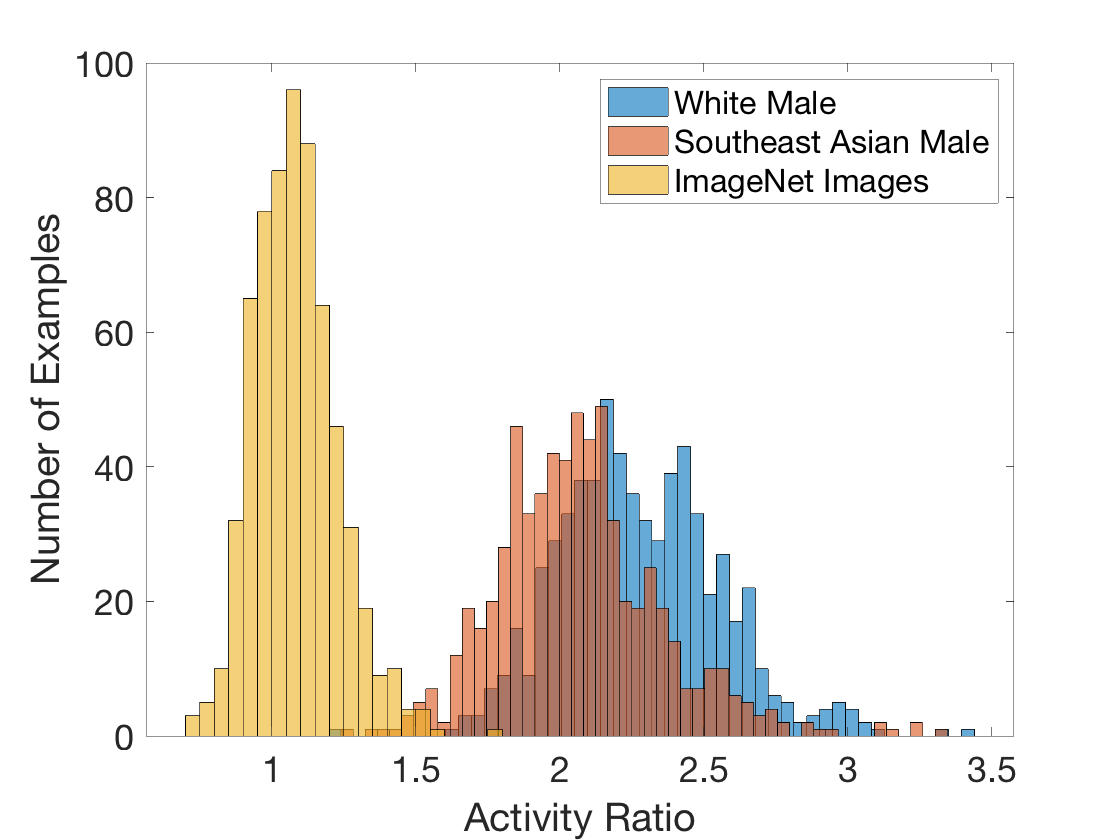}}
    \subfigure[Southeast Asian / F]{\includegraphics[width=2.8cm]{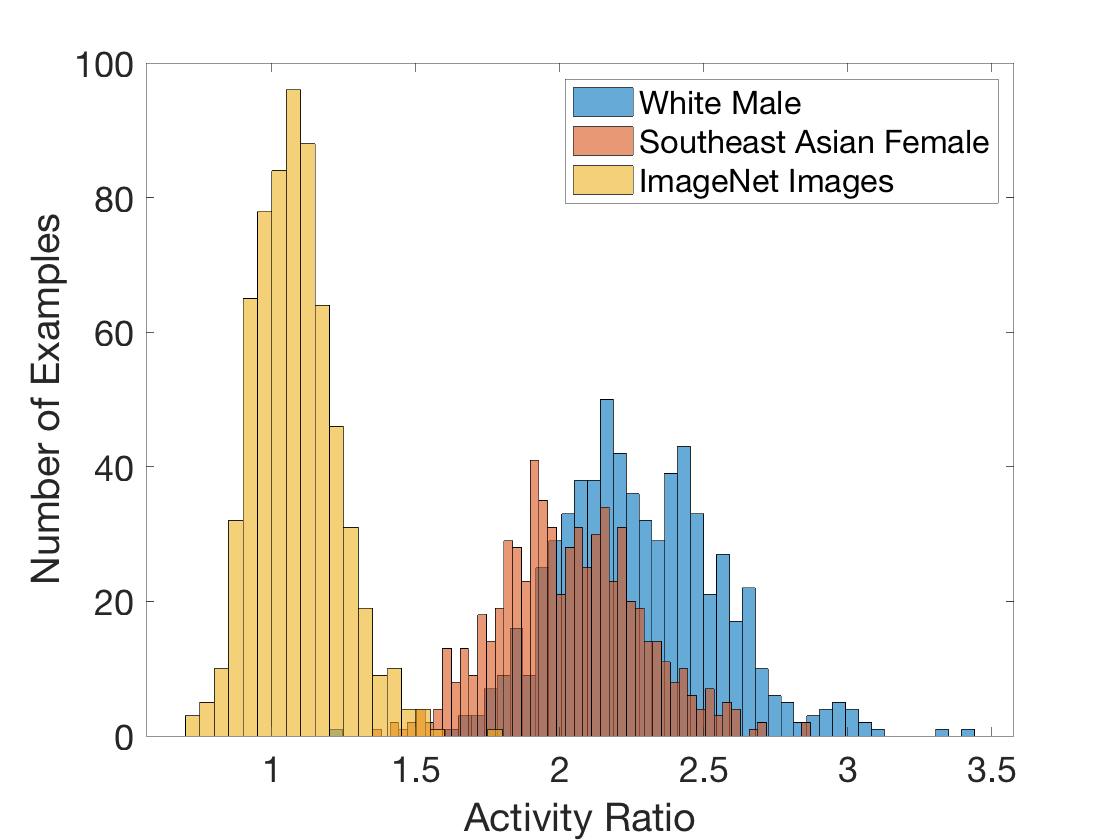}}
  \caption{Histogram of the ratio of activity shown by the FFA and IT neurons for White Male faces from FairFace compared to various other ethnic categories and ImageNet images.  The MDCA model was trained on 800 White Male faces only.  Even with the minor distribution shift between categories, the faces and objects can still be easily separated by an activity threshold around 1.4.}
\vspace{-0.2cm}
\label{fig:histo}
\end{figure*}
While our model built in a biologically-plausible manner can explain a multitude of biological phenomena, our final experiment elucidates strategies for building robust category level classifiers (in our example, face or not face) that exhibit less bias.  Furthermore, this is remarkably achieved via unsupervised learning, competition, and a simple threshold rather than through a supervised objective. 

A survey of nearly 200 facial recognition systems, performed by the National Institute of Standards and Technology, showed false identifications of Asian, African American, and Native American faces at higher rates than Caucasian faces. Non-Caucasian faces were falsely identified from 10 to 100 times more often, depending on the algorithm \cite{NIST2019frvt}. Others uncovered severely worse performance of commercially-used face analysis systems on darker-skinned females \cite{buolamwini2018gender, klare2012demographics}. 
The under performance in such systems was likely caused by representation bias in the training data and the evaluation benchmark datasets failed to discover and penalize them \cite{suresh2019framework}.  Reweighting instances within the dataset \cite{jiang2020identifying} and resampling or balancing the datasets \cite{de2019does} can improve performance, however, these underlying biases were still present. 



These enormous failures, up to 35\% in face detection algorithms on darker-skinned women, were attributed to datasets heavily dominated by male and white examples \cite{lohr2018facial}.  On the other hand, human beings are of often presented with heavily biased, long tailed training data but are still able to learn off relatively few examples \cite{lake2015human,gerken2006}.  


In response, we challenged our MDCA model to generalize in the task of face detection out of distribution.  We compared our framework to a standard deep learning CNN model and fine-tuned, off the shelf modles (ResNet-50 \cite{he2016deep}, VGG16\cite{simonyan2014very}) trained with data from the FairFace dataset \cite{karkkainen2019fairface} and ImageNet. The FairFace dataset contains images of people from seven ethnicity groups, across a wide range of variations. 

Specifically, for the custom CNN, we built a 3-layer CNN binary classifier (face/not face) that matches the architecture, size, and parameters of a single pathway of our MDCA model.  We trained with a biased and unbalanced dataset consisting of 800 White-Male faces and 10,000 ImageNet images.  Our in-distribution test set contained 200 White-Male faces and 1,000 ImageNet images. 

We trained the CNN for 35 epochs, fine-tuned ResNet50 and VGG16 for 35 epochs, and pre-trained the MDCA pathways with the same images and for an approximately equivalent number of image impressions, i.e. 35 epochs x 10,800 images = 378,000 image impressions, and we trained the MDCA pathway for 375,000 timesteps.  The CNN is trained with a supervised loss, whereas the MDCA seeks to minimize reconstruction loss, and classification of face/non-face is achieved through a simple 1.4 threshold on the ratio of activity as shown in Figure \ref{fig:histogram}.

Our results were astounding.  For the in-distribution test set, the custom CNN, VGG16, ResNet50, and MDCA 
performed very well as expected, 97.17\%, 99.83\%, 99.67\%, and 98.25\% respectively.  However, as noted in previous literature,  deep learning models struggle to generalize its understandings in face detection of one ethnicity and gender to other categories, failing in over 36\% of the cases on Black males in the custom CNN. MDCA, on the other hand, was capable of detecting faces of every ethnicity and gender with nearly perfect accuracy in all categories. The results are presented in Table \ref{tab:bias_acc_comparison} and Figure \ref{fig:histo}. 

Interestingly, our model definitely still reflects the bias of the training data as shown in Figure \ref{fig:bias_res1} and \ref{fig:bias_res2}.  One can see that the FFA can only contribute white male faces to the reconstruction, yet due to the holistic, coarse-to-fine nature of the model, the finer details can be captured and incorporated by the lower layers of the model.  In summary, the dynamic and competitive nature of our approach allows one part of the model to overcome the bias exhibited by a different part of the model.

\begin{table}[tbh]
\centering
\footnotesize
\setlength\tabcolsep{3.5pt}
\begin{tabular}{llccccc}
\hline
\multicolumn{2}{l}{\textbf{Ethnicity/Gender}} & \multicolumn{1}{l}{\textbf{\#img}} & \multicolumn{1}{l}{\textbf{CNN}} & \multicolumn{1}{l}{\textbf{ResNet50}} & \multicolumn{1}{l}{\textbf{VGG16}} & \multicolumn{1}{l}{\textbf{MDCA}} \\ \hline
\multicolumn{2}{l}{Black/F} & 757 & 73.84 & 97.09 & 95.64 & \textbf{99.33} \\
\multicolumn{2}{l}{Black/M} & 799 & 63.2 & 97.12 & 94.49 & \textbf{99.87} \\ \hline
\multicolumn{2}{l}{East-Asian/F} & 773 & 83.31 & 95.60 & 95.21 & \textbf{99.61} \\
\multicolumn{2}{l}{East-Asian/M} & 777 & 77.09 & 96.14 & 96.78 & \textbf{99.87} \\ \hline
\multicolumn{2}{l}{Indian/F} & 763 & 88.33 & 96.59 & 96.20 & \textbf{100} \\
\multicolumn{2}{l}{Indian/M} & 753 & 88.34 & 95.88 & 94.95 & \textbf{100} \\ \hline
\multicolumn{2}{l}{Latino-Hispanic/F} & 830 & 86.86 & 96.63 & 98.07 & \textbf{99.63} \\
\multicolumn{2}{l}{Latino-Hispanic/M} & 793 & 83.98 & 95.88 & 96.47 & \textbf{100} \\ \hline
\multicolumn{2}{l}{Middle-Eastern/F} & 396 & 83.83 & 94.19 & 94.95 & \textbf{100} \\
\multicolumn{2}{l}{Middle-Eastern/M} & 813 & 82.41 & 96.06 & 95.82 & \textbf{99.38} \\ \hline
\multicolumn{2}{l}{Southeast-Asian/F} & 680 & 85.00 & 97.21 & 95.74 & \textbf{99.85} \\
\multicolumn{2}{l}{Southeast-Asian/M} & 735 & 81.49 & 98.23 & 97.01 & \textbf{99.59} \\ \hline
\end{tabular}
\caption{Classification accuracy on different ethnicity categories and genders, a comparison among 4 different models, custom CNN, ResNet50, VGG16, and our model, MDCA.}
\label{tab:bias_acc_comparison}
\vspace{-0.3cm}
\end{table}

\section{Conclusion and Future Work}
In conclusion, we created a computational model that incorporates important thematics observed in the brain, selectivity through competition,  dedicated pathways, holistic/coarse-to-fine processing, and top-down feedback.   We demonstrate that these neural mechanisms provide the foundation of a novel, robust classification framework that rivals traditional supervised learning in computer vision. In the example of machine learning bias, we convincingly demonstrate that a self-supervised model rooted in competition can significantly out perform a supervised deep learning model.  Furthermore, we note that there are many other computer vision applications that are enabled by such a generative framework shown in our supplemental material.  


{\small
\bibliographystyle{ieee_fullname}
\bibliography{egbib}
}
\end{document}